%% file: main.tex
\def\year{2021}\relax
\documentclass[letterpaper]{article} 
\usepackage{aaai21}  
\usepackage{times}  
\usepackage{helvet} 
\usepackage{courier}  
\usepackage[hyphens]{url}  
\usepackage{graphicx} 
\usepackage{natbib}  
\usepackage{caption} 
\input{packages}

\title{Robust Finite-State Controllers for Uncertain POMDPs}%

\author{
	Murat Cubuktepe$^1$,
	Nils Jansen$^2$,
	Sebastian Junges$^3$,\\
	Ahmadreza Marandi$^4$,
	Marnix Suilen$^2$,
	Ufuk Topcu$^1$\\%
}%
\affiliations{
	\textsuperscript{\rm 1}Department of Aerospace Engineering and Engineering Mechanics, University of Texas at Austin, USA\\
	\textsuperscript{\rm 2}Department of Software Science, Radboud University, The Netherlands\\
	\textsuperscript{\rm 3}Department of Electrical Engineering and Computer Sciences, University of California, Berkeley, USA\\
	\textsuperscript{\rm 4}Department of Industrial Engineering and Innovation Sciences, Eindhoven University of Technology, The Netherlands\\%
}%

\input{macros}

\begin{document}
	
	\maketitle
	
	\input{abstract}

	\input{introduction}

	\input{preliminaries}
	
	\input{problem}
	
		\input{uncertainty}
	
	\input{convexification}

	\input{experiments}

	\input{conclusion}

%
	\section{Acknowledgments}
	
	M. Cubuktepe and U. Topcu were supported in part by the grants ARL ACC-APG-RTP W911NF, NASA 80NSSC19K0209, and AFRL FA9550-19-1-0169.
	N. Jansen and M. Suilen were supported in part by the grants NWO OCENW.KLEIN.187: ``Provably Correct Policies for Uncertain Partially Observable Markov Decision Processes'', NWA.1160.18.238: ``PrimaVera'',  and NWA.BIGDATA.2019.002: ``EXoDuS - EXplainable Data Science''.
	S. Junges was supported in part by NSF grants 1545126 (VeHICaL) and 1646208, by the DARPA Assured Autonomy program, by Berkeley Deep Drive, and by Toyota under the iCyPhy center.
	
	%
	\bibliography{literature}
	
	

\end{document}

%% file: packages.tex
\usepackage{amsmath,amssymb,amsfonts,amsthm}       
\usepackage{mathtools}
\usepackage{microtype}
\usepackage{todonotes}
\usepackage{colonequals}
\usepackage{pgfplots}
\usetikzlibrary{patterns,arrows,arrows.meta,calc,shapes,shadows,decorations.pathmorphing,decorations.pathreplacing,automata,shapes.multipart,positioning,shapes.geometric,fit,circuits,trees,shapes.gates.logic.US,fit}
\usetikzlibrary{backgrounds,scopes}
\usepackage{multirow}
\usepackage{algorithm}
\usepackage{algpseudocode}
\usepackage{paralist}
\usepackage{scalerel,stackengine}

\usepackage{subcaption}
\usepackage{nicefrac}
\usepackage{tikz} 
\usepackage{mdframed}

\usetikzlibrary{arrows,decorations.pathmorphing,positioning,fit,trees,shapes,shadows,automata,calc} 

\theoremstyle{definition}
\newtheorem{definition}{Definition}

\theoremstyle{remark}

\usepackage{scalerel,stackengine}


%% file: macros.tex
\newcommand{\rew}{\ensuremath{R}}

\newcommand{\expRewProp}[2]{\ensuremath{\EV_{\geq #1}(\finally #2)}}

\newcommand{\finally}{\lozenge}

\newcommand{\R}{\mathbb{R}}

\newcommand{\EV}{\ensuremath{\mathbb{E}}}

\newcommand{\Distr}{\mathit{Distr}}

\newcommand{\distDom}{X}

\newcommand{\sinit}{s_{\mathit{I}}} 

\newcommand{\probumdp}{\mathcal{P}}
\newcommand{\intervals}{\mathbb{I}}

\newcommand{\sched}{\ensuremath{\sigma}}

\newcommand{\Act}{\ensuremath{\mathit{A}}}

\newcommand{\act}{\ensuremath{\alpha}}

\DeclareMathAlphabet{\mathpzc}{OT1}{pzc}{m}{it}
\def\presuper#1#2%
  {\mathop{}%
   \mathopen{\vphantom{#2}}^{#1}%
   \kern-\scriptspace%
   #2}

\renewcommand{\paragraph}[1]{\par\smallskip\noindent\textbf{#1}}

\newcommand{\ObsSym}{{Z}}
\newcommand{\ObsFun}{{O}}
\newcommand{\obs}{\ensuremath{z}}

\newcommand{\uPomdpInit}[1][]{\pomdp{#1}=(S{#1},\sinit{#1},\Act,\intervals,\probumdp{#1},\ObsSym{#1},\ObsFun{#1},\rew)}
\newcommand{\pomdp}{\mathcal{M}}

\newcommand{\states}{\ensuremath{S}}

\newcommand{\osched}{\ensuremath{\mathit{\sigma}}}

\newcommand\pig[1]{\scalerel*[5pt]{\big#1}{%
		\ensurestackMath{\addstackgap[1.5pt]{\big#1}}}}
	

%% file: abstract.tex
\begin{abstract}
Uncertain partially observable Markov decision processes (uPOMDPs) allow the probabilistic transition and observation functions of standard POMDPs to belong to a so-called uncertainty set.
Such uncertainty, referred to as epistemic uncertainty, captures uncountable sets of probability distributions caused by, for instance, a lack of data available.
We develop an algorithm to compute finite-memory policies for uPOMDPs that robustly satisfy specifications against any admissible distribution.
In general, computing such policies is theoretically and practically intractable.
We provide an efficient solution to this problem in four steps.
(1) We state the underlying problem as a nonconvex optimization problem with infinitely many constraints. 
(2) A dedicated dualization scheme yields a dual problem that is still nonconvex but has finitely many constraints. 
(3) We linearize this dual problem 
and (4) solve the resulting finite linear program to obtain locally optimal solutions to the original problem.
The resulting problem formulation is exponentially smaller than those resulting from existing methods.
We demonstrate the applicability of our algorithm using large instances of an aircraft collision-avoidance scenario and a novel spacecraft motion planning case study.
\end{abstract}

%% file: introduction.tex
\section{Introduction}
\label{sec:introduction}
In sequential decision making, one has to account for various sources of uncertainty.
Examples for such sources are lossy long-range communication channels with a spacecraft orbiting the earth, or the expected responses of a human operator in a decision support system~\cite{kochenderfer2015decision}.
Moreover, sensor limitations may lead to imperfect information of the current state of the system.
The standard models to reason about decision-making under uncertainty and imperfect information are partially observable Markov decision processes (POMDPs)~\cite{kaelbling1998planning}.

The likelihood of uncertain events, like a message loss in communication channels or of specific responses by human operators, is often only estimated from data.
Yet, such likelihoods enter a POMDP model as concrete probabilities. 
POMDPs, in fact, \emph{require} precise transition and observation probabilities.
Uncertain POMDPs (uPOMDPs) remove this assumption by incorporating uncertainty sets of probabilities~\cite{burns2007sampling}.
These sets are part of the transition and observation model in uPOMDPs and take the form of, for example, probability intervals or likelihood functions~\cite{givan2000bounded,DBLP:journals/ior/NilimG05}.
Such uncertainty is commonly called \emph{epistemic}, which is induced by a lack of precise knowledge as opposed to \emph{aleatoric} uncertainty, which is due to stochasticity.

\paragraph{Motivating example.} Consider a robust spacecraft motion planning system which serves as decision support for a human operator.
This system delivers advice on switching to a different orbit or avoiding close encounters with other objects in space.
Existing POMDP models assume that a fixed probability captures the responsiveness of the human~\cite{kochenderfer2015decision}.
However, a single probability may not faithfully capture the responsiveness of multiple operators.
Instead of using a single probability, we use bounds on the actual value of this probability and obtain a uPOMDP model. 


A policy for the decision support system necessarily considers all potential probability distributions and may need to predict the current location of the spacecraft based on its past trajectory. 
Therefore, it requires \emph{robustness} against the uncertainty, and \emph{memory} to store past (observation) data. 
%
%
%

\paragraph{Problem formulation.} The general problem we address is:
\begin{mdframed}[backgroundcolor=gray!30]
	For a uPOMDP, compute a policy that is robust against all possible probabilities from the uncertainty set.
\end{mdframed}%
%
Even without uncertainties, the problem is undecidable~\cite{MadaniHC99}. Most research has focused on either belief-based approaches or on computing finite state controllers (FSCs)~\cite{meuleau1999solving,amato2010optimizing}.
Our approach extends the latter and computes \emph{robust} FSCs representing robust policies.




\subsection{Contribution and Approach}
We develop a novel solution for efficiently computing robust policies for uPOMDPs using robust convex optimization techniques.
The method solves uPOMDPs with hundreds of thousands of states, which is out of reach for the state-of-the-art.
\looseness=-1

\paragraph{Finite memory.}
The first step of our method is to obtain a tractable representation of memory.
As a concise representation of uPOMDP policies with \emph{finite memory}, we utilize FSCs. 
First, we build the product of the FSC memory and the uPOMDP.
Computing a policy with finite memory for the original uPOMDP is equivalent to computing a \emph{memoryless} policy on the product~\cite{junges2018finite}. 
While the problem complexity remains NP-hard~\cite{VlassisLB12}, the problem size grows polynomially in the size of the FSC.
Existing methods cannot cope with that blow-up, as our experiments demonstrate.

\paragraph{Semi-infinite problem.}
We state the uPOMDP problem as a semi-infinite nonconvex optimization problem with infinitely many constraints. 
A tailored \emph{dualization} scheme translates this problem into a finite optimization problem with  a linear increase in the number of variables.
Combining this dualization with a linear-time transformation of the uPOMDP to a so-called \emph{simple uPOMDP}~\cite{junges2018finite} ensures \emph{exact solutions} to the original problem.
The exact solutions and the moderate increase in the problem size contrasts with an over-approximative solution computed using an exponentially larger encoding proposed in~\cite{suilen2020robust}.

\paragraph{Finite nonconvex problem.}
The resulting finite optimization problem is still nonconvex and infeasible to solve exactly~\cite{alizadeh2003second,lobo1998applications}. 
We adapt a \emph{sequential convex programming} (SCP) scheme~\cite{mao2018successive} that iteratively linearizes the problem around previous solutions. 
Three extensions help to alleviate the errors introduced by the linearization, rendering the method sound for the uPOMDP problem.

\paragraph{Numerical experiments.}
We demonstrate the applicability of the approach using two case studies.
Notably, we introduce a novel robust spacecraft motion planning scenario.
We show that our method scales to significantly larger models than~\cite{suilen2020robust}.
This scalability advantage allows more precise models and adding memory to the policies.

\subsection{Related work} 
Uncertain MDPs with full observability have been extensively studied, see for instance~\cite{DBLP:conf/cdc/WolffTM12,DBLP:conf/cav/PuggelliLSS13,DBLP:conf/qest/HahnHHLT17}.
%
For uPOMDPs, \cite{suilen2020robust} is also based on convex optimization.
However, their resulting optimization problems are exponentially larger than ours, and they only consider memoryless policies. 
\cite{burns2007sampling} utilizes sampling-based methods and~\cite{DBLP:conf/icml/Osogami15} employs a robust value iteration on the belief space of the uPOMDP. 
\cite{ahmadi2018verification} uses sum-of-squares optimization to verify uPOMDPs against temporal logic specifications.
The work in \cite{itoh2007} assume distributions over the probability values of the uncertainty set.
Finally,~\cite{DBLP:conf/cdc/ChamieM18} considers a convexified belief space and computes a policy that is robust over this space. 

%% file: preliminaries.tex
\section{Background and Formal Problem}
\label{sec:preliminaries}

\paragraph{Uncertain POMDPs.}
The set of all probability distributions over a finite or countably infinite set $\distDom$ is $\Distr(\distDom)$.
%


\begin{definition}[uPOMDP]
\label{def:umdp}
  An \emph{uncertain partially observable Markov decision process (uPOMDP)} is a tuple $\uPomdpInit$ with a finite set $\states$ of states, an initial state $\sinit \in \states$, a finite set $\Act$ of actions,
  a set  $\intervals$ of probability intervals, an \emph{uncertain transition function} $\probumdp \colon \states \times \Act \times \states \to \intervals$, a finite set $\ObsSym$ of \emph{observations}, an \emph{uncertain observation function} $\ObsFun\colon\states\times \ObsSym \to\intervals$, and a reward function $R \colon \states \times \Act \to \R_{\geq 0}$.
\end{definition}
%
%
%
%
\noindent \emph{Nominal} probabilities are point intervals where the upper and lower bounds coincide.
If all probabilities are nominal, the model is a (nominal) POMDP. 
We see uPOMDPs as sets of nominal POMDPs that vary only in their transition function.
For a transition  function $P \colon \states \times \Act \times S \to \mathbb{R}$, we write $P \in \probumdp$ if for all $s,s' \in \states$ and $\alpha \in \Act$ we have $P(s,\act,s') \in \probumdp(s,\act,s')$ and $P(s, \act, \cdot)$ is a probability distribution over $S$.
Finally, a fully observable uPOMDP where each state has unique observations is an \emph{uncertain MDP}.

\paragraph{Policies.}
  An \emph{observation-based policy} $\osched\colon (Z \times \Act)^*\times Z\rightarrow\Distr(\Act)$ for a uPOMDP maps a \emph{trace}, i.e., a sequence of observations and actions, to a distribution over actions. 
A \emph{finite-state controller} (FSC) consists of a finite set of memory states and two functions. 
The \emph{action mapping} $\gamma(n,\obs)$ takes an FSC memory state $n$ and an observation $\obs$, and returns a distribution over uPOMDP actions.
To change a memory state, the \emph{memory update} $\eta(n,\obs,\act)$ returns a distribution over memory states and depends on the action $\act$ selected by $\gamma$.
An FSC induces an observation-based policy by following a joint execution of these functions upon a trace of the uPOMDP.
An FSC is \emph{memoryless} if there is a single memory state. Such FSCs encode policies $\osched\colon Z \rightarrow\Distr(\Act)$.
   
\paragraph{Specifications.}   
We constrain the undiscounted expected reward (the value) of a policy for a uPOMDP using \emph{specifications}:
For a POMDP $\pomdp$ and a set of goal states $G$
the specification $\expRewProp{\kappa}{G}$ states that the expected reward before reaching $G$ is at least $\kappa$.
For brevity, we require that the POMDP has no dead-ends, i.e., that under every policy, we eventually reach $G$.
Reachability specifications to a subset of $G$ and discounted rewards are special cases~\cite{Put94}.
%
%
%

\paragraph{Satisfying specifications.}
A policy $\osched$ satisfies a specification $\varphi=\expRewProp{\kappa}{G}$, if the expected reward to reach $G$ induced by $\osched$ is at least $\kappa$.
POMDP $\pomdp[P]$ denotes the \emph{instantiated} uPOMDP $\pomdp$ with a fixed transition function $P \in \mathcal{P}$.
A policy \emph{robustly satisfies} $\varphi$ for the uPOMDP $\pomdp$, if it does so for all $\pomdp[P]$ with $P\in\probumdp$. Thus, a (robust) policy for uPOMDPs accounts for all possible instantiations $P \in \probumdp$.

\paragraph{Formal problem statement.}
Given a uPOMDP $\pomdp$ and a specification $\varphi$,
 compute an FSC that yields an observation-based policy $\osched$ which robustly satisfies $\varphi$ for $\pomdp$.\smallskip

\paragraph{Assumptions.}
We compute the product of the memory update of an FSC and the uPOMDP and reduce the problem do computing policies without memory~\cite{junges2018finite}.
Therefore, we focus in the following on memoryless policies and show the impact of finite memory in our examples.

%

%

	For the correctness of our method, instantiations are not allowed to remove transitions. That is, for any interval we either require the lower bound to be strictly greater than $0$ such that the transition \emph{exists in all instantiations}, or both upper and lower bound to be equal to $0$ such that the transition \emph{does not exist} in all instantiations.
	This assumption is standard and for instance used in~\cite{wiesemann2013robust}.



\paragraph{Outline.} 
In Sect.~\ref{sec:problem}, we equivalently formulate our problem as semi-infinite nonlinear problem (NLP).
In Sect.~\ref{sec:uncertainty}, we translate this semi-infinite NLP can be transformed into a finite NLP.
Then, in Sect.~\ref{sec:scp}, we linearize the finite NLP into a finite linear program, and solve the resulting linear program (LP), inspired by so-called sequential convex programming methods. 
In Sect.~\ref{sec:experiments}, we evaluate our method.

%% file: problem.tex
%


\section{Optimization Problem for uPOMDPs}
\label{sec:problem}
In this section, we reformulate our problem statement as a semi-infinite nonconvex optimization problem, with finitely many variables but infinitely many constraints.
We utilize simple uPOMDPs, defined below, as the dualization discussed in Sect.~\ref{sec:uncertainty} is exact (only) for simple uPOMDPs. 

We adopt a small extension of (PO)MDPs in which only a subset of the actions are available in a state, i.e., the transition function should be interpreted as a partial function.  We denote the set of actions at state $s$ by $\Act(s)$. We ensure that states sharing an observation share the set of available actions.
Moreover, we translate the observation function to be deterministic without uncertainty, i.e., of the form $\ObsFun\colon\states\rightarrow \ObsSym$, by expanding the state space~\cite{ChatterjeeCGK16}.
\begin{definition}[Binary/Simple uPOMDP]
	A uPOMDP is \emph{binary}, if $|\Act(s)|\leq 2$ for all $s \in S.$ A binary uPOMDP is \emph{simple} if for all $s \in S$, the following is true:
	\begin{align*}
	|\Act(s)|=2 \;\text{implies}\; \forall \act \in \Act(s), \;\exists s' \in S, \;\probumdp(s,\act,s') =1,
	\end{align*}
\end{definition}%
Simple uPOMDPs differentiate the states with \emph{action} choices and \emph{uncertain} outcomes. 
All uPOMDPs can be transformed into simple uPOMDPs. 
We refer to~\cite{junges2018finite} for a transformation.
This transformation preserves the optimal expected reward of a uPOMDP.

Let $S_{\mathrm{a}}$ denote the states with action choices, and $S_{\mathrm{u}}$ denote the states with uncertain outcomes.
We now introduce the optimization problem with the nonnegative reward variables $\{r_s \geq 0\; | \; s \in S\}$ denoting the expected reward before reaching goal set $G$ from state s, and positive variables $\{\sched_{s,\act}> 0 \; | \; s \in S, \act \in \Act(s)\}$ denoting the probability of taking an action $\act$ in a state $s$ for the policy.
    Note that we only consider policies where for all states $s$ and actions $\alpha$ it holds that $\sched_{s,\act} > 0$, such that applying the policy to the uPOMDP does not change the underlying graph.%
    \begin{flalign}
&\begin{aligned}
\text{maximize} \quad   & r_{s_I}
\end{aligned} \label{NLP:obj}\\
&\begin{aligned}
\text{subject to} \quad
\,\,  r_{s_I} \geq \kappa, \quad \forall s \in G, \quad & r_s = 0,
\end{aligned}  \label{NLP:constraint:2}\\
%
&\begin{aligned} 
\forall s \in S, \quad  & \sum\nolimits_{\act \in \Act(s)} \sched_{s,\act} = 1,
\end{aligned} \label{NLP:constraint:3}\\
&\begin{aligned} 
\forall s, s' \in S \text{ s.t.\ } \ObsFun(s) = \ObsFun(s'), \forall \alpha \in \Act(s),\; \sigma_{s,\alpha} = \sched_{s',\alpha},
\end{aligned}\label{NLP:constraint:4}
\raisetag{0.8\baselineskip}\\
&\begin{aligned} 
&\!\forall s \in S_{\mathrm{a}}, \,r_s \leq \!\!\!\sum_{\act \in \Act(s)}  \! \! \! \sched_{s,\act} \!\cdot \!\pig(\rew(s,\act)\! + \!\! \sum_{s' \in S} \! \probumdp(s,\act,s')\cdot r_{s'}\pig),
\end{aligned} \label{NLP:simple-nondet}\raisetag{0.80\baselineskip}\\
&\begin{aligned} 
&\forall s \in S_{\mathrm{u}},\, \forall P \in \mathcal{P},\;  
\, r_s \leq R(s)+\sum_{s' \in S} P(s,s')\cdot r_{s'}.
\end{aligned}\label{NLP:simple-uncertain} \raisetag{1.6\baselineskip}
\end{flalign}%
The objective is to maximize the expected reward $r_{s_I}$ at the initial state.
The constraint~\eqref{NLP:constraint:2} encodes the specification requirement and assigns the expected reward to $0$ in the states of goal set $G$.
We ensure that the policy is a valid probability distribution in each state by~\eqref{NLP:constraint:3}.
Next,~\eqref{NLP:constraint:4} ensures that the policy is observation-based.
We encode the computation of expected rewards for states with action choices by~\eqref{NLP:simple-nondet} and with uncertain outcomes by~\eqref{NLP:simple-uncertain}.
We omit denoting the unique actions in the transition function $P(s,s')$ and reward function $R(s)$ in~\eqref{NLP:simple-uncertain} for states with uncertain outcomes.

Let us consider some properties of the optimization problem.
First, the functions in~\eqref{NLP:simple-nondet} are \emph{quadratic}. 
Essentially, the policy variables $\sched_{s,\act}$ are multiplied with the reward variables $r_{s}$. In general, these constraints are \emph{nonconvex}, and we later \emph{linearize} them.
Second, the values of the transition probabilities $P(s,s')$ for $s, s' \in S_{\textrm{u}}$ in~\eqref{NLP:simple-uncertain}  belong to continuous intervals.
Therefore, there are infinitely many constraints over a finite set of reward variables. These constraints are similar to the LP formulation for MDPs~\cite{Put94}, and are affine; there are no policy variables.
%

%% file: uncertainty.tex
\section{Dualization of Semi-Infinite Constraints}
\label{sec:uncertainty}
In this section, we transform the semi-infinite optimization problem into a finite optimization problem using \emph{dualization} for simple uPOMDPs.
Specifically, we translate the semi-infinite constraints from~\eqref{NLP:simple-uncertain} into finitely many constraints.

\subsection{Dualization for Robust Linear Programs}
\label{sec:duality}
We summarize robust LPs with polytopic uncertainty~\cite{bertsimas2011theory,ben2009robust}.
The idea for solving robust LPs with such uncertainty is essential in our approach.%
\paragraph{Robust LPs.}
	\noindent A robust LP with the variable $x \in \R^n$ is 
	\begin{align*}
	&\text{minimize} \quad c^\top x\\
	&\text{subject to} \quad (d+u)^\top x \leq e \quad \forall u \in \mathcal{U},
	\end{align*}
	where $c, d \in \R^n$, and $e \in \R$ are given vectors, $u \in \R^n$ is the uncertain parameter, and $\mathcal{U}$ is the \emph{uncertainty set}. 
	We assume that the set $\mathcal{U}$ is a convex polytope, i.e., that it is an $n$-dimensional shape defined by the linear inequalities  $Cu +g\geq 0$ for $C \in \R^{m \times n}$ and $g \in \R^m$. 

\paragraph{Duality.}
For simplicity, we explain the idea on a single robust inequality.
The idea can be generalized to multiple robust inequalities.
With $\mathcal{U}$ defined by the linear inequalities above, \emph{duality} can be used to obtain a tractable solution to the robust LP.
Specifically, we write the Lagrangian for the maximization problem over $u^\top x$ with the dual variable $\mu\geq 0$ as
\begin{align*}
L(u,\mu)= u^\top x + \mu^{\top}  (C u + g).
\end{align*}
By taking the supremum over $u$, we obtain
\begin{align*}
\underset{u \in \mathcal{U}}{\text{sup}} \;\;L(u,\mu)=   \begin{cases}
\infty & \text{if } C^{\top}\mu + x \neq 0, \\
\mu^\top g & \text{if } C^{\top}\mu + x = 0.
\end{cases}
\end{align*}
All inequalities are linear which implies \emph{strong duality}, i.e.,
\begin{align*}
\underset{u \in \mathcal{U}}{\text{sup}}\;\; u^\top x = \underset{\mu \geq 0}{\text{inf}}\;\; \lbrace \mu^\top g \;  |  \; C^{\top}\mu + x = 0\rbrace.
\end{align*}
Since these optimization problems are linear, we know that the optimal value is attained at a feasible solution. 
Therefore, we can replace $\text{sup}$ and $\text{inf}$ with $\text{max}$ and $\text{min}$.
The semi-infinite inequality with polytopic uncertainty is equivalent to the following linear constraints%
\begin{align*}
d^{\top}x+\mu^{\top}g \leq e, \quad C^\top\mu + x =0, \quad \mu \geq 0.
\end{align*}%
\subsection{Dualization for Simple POMDPs}
We now describe the dualization step that we use to obtain a finite optimization problem for simple POMDPs.
\label{sec:simple_dual}
\paragraph{Uncertainty polytopes.}
\label{sec:poly}
First, we formalize the uncertainty sets for simple uPOMDPs. 
For state $s\in S_\mathrm{u}$, we assume that there are $|S|$ intervals $[a_s, b_s]\in\intervals$, $s \in S$ that describe the uncertain probability of transitioning to a successor state of $s$.
This setting is equivalent to~\cite{suilen2020robust}.

The following constraints ensure that the uncertain transition function $P$ is well-defined at a state $s$:
\begin{align*}
\forall s' \in S, \quad  a_{s,s'} \leq u_{s,s'} \leq b_{s,s'}, \,\,\,\, \sum\nolimits_{s' \in S} u_{s,s'} = 1.
\end{align*}
We denote these constraints in matrix form: $C_su+g_s\geq 0$. 
%

\paragraph{Dualization.}
After obtaining the matrices $C_s$ and vectors $g_s$ characterizing uncertainty sets for each state, we directly use dualization to transform the inequalities in~\eqref{NLP:simple-uncertain} into \begin{align}
	\hspace{-0.1cm}\forall s \in S_{\mathrm{u}},\, r_s \leq R(s)+\mu^{\top}_s g_s,\,
	 C^\top_s\mu_s+q=0, \,\mu_s \geq 0,\label{NLP:simple-uncertain-finite}
\end{align}%
where $\mu_s\in \R^{|S|}$ is the dual variable of the constraint $C_su+g_s\geq 0$ and $q$ is an $|S|-$dimensional vector denoting the set of reward variables for each state $s \in S$. 
After this step, the problem with the objective~\eqref{NLP:obj} and the constraints \eqref{NLP:constraint:2}--\eqref{NLP:simple-nondet} and~\eqref{NLP:simple-uncertain-finite} has finitely many variables and constraints.

In the worst-case, the number of dual variables in~\eqref{NLP:simple-uncertain-finite} is $|S|\cdot|S_u|$, and the number of constraints is $2\cdot|S|\cdot|S_u|+|S_u|$, which only yields a factor $|S|$ of increase in the number of variables compared to the aforementioned semi-infinite optimization problem.
In practice, most of the values of $a_{s,s'}$, and $b_{s,s'}$ are $0$, meaning there are few successor states for each state $s$, and we exploit this structure in our numerical examples. 
\looseness=-1

%% file: convexification.tex
\section{Linearizing the Finite Nonconvex Problem}
\label{sec:scp}
In this section, we discuss our algorithm to solve the (finite but nonconvex) dual problem from the previous section.
Our method is based on a \emph{sequential convex programming} (SCP) method~\cite{yuan2015recent,mao2018successive}.
SCP iteratively computes a locally optimal solution to the dual problem from Section~\ref{sec:simple_dual} by approximating it as an LP.
In every iteration, this approximation is obtained by \emph{linearizing} the quadratic functions around a previous solution.
The resulting LP does not necessarily have a feasible solution, and the optimal solution does not necessarily correspond to a feasible solution of the original dual problem.
We alleviate this drawback by adding three components to our approach.


\paragraph{Linearizing.}
To linearize the dual problem, we only need to linearize the quadratic functions in~\eqref{NLP:simple-nondet}, repeated below.
\begin{align*} 
\!\forall s \in S_{\mathrm{a}}, r_s \leq \!\!\sum_{\act \in \Act(s)} \! \! \!  \sched_{s,\act} \!\cdot \!\Big(\rew(s,\act)\! +  \! \sum_{s' \in S} \! \probumdp(s,\act,s')\cdot r_{s'}\Big)\raisetag{0.7\baselineskip}
\end{align*}
The constraint is quadratic in $r_{s}$ and  $\sched_{s,\act}$. 
For compact notation, let $h(s,\act,s')$
be the quadratic function above for a given $s \in S_\textrm{a}, s' \in S$ and $\act \in \Act(s)$:
\begin{align*}
h(s,\act,s')=\sched_{s,\act} \cdot\probumdp(s,\act,s')\cdot r_{s'}
\end{align*}
and let $
d=\probumdp(s,\act,s'), y=\sched_{s,\act}, \text{ and } z=r_{s'}$.
Let $\langle\hat{y}, \hat{z}\rangle$ denote an arbitrary assignment to $y$ and $z$.
We linearize $h(s,\act,s')$ around $\langle\hat{y}, \hat{z}\rangle$ as
\begin{align*}
h_{\textrm{aff}}(s,\act,s') &=
 d\cdot\big(\hat{y}\cdot\hat{z}+\hat{y}\cdot(z-\hat{z})+\hat{z}\cdot(y-\hat{y})\big)\\
 & = d\cdot(\hat{y}\cdot z+\hat{z}\cdot y-\hat{z}\cdot \hat{y}).
\end{align*} 
The resulting function $h_{\textrm{aff}}(s,\act,s')$ is affine in $y$ and $z$ ($r_{s}$ and $\sched_{s,\act}$).
After the linearization step, we replace~\eqref{NLP:simple-nondet} by 
\begin{align}
\!\forall s \in S_{\mathrm{a}}, r_s \leq \!\!\!\sum_{\act \in \Act(s)}  \! \! \! \sched_{s,\act} \cdot \Big(\rew(s,\act)\! +  \sum_{s' \in S}h_{\textrm{aff}}(s,\act,s')\Big)\raisetag{0.65\baselineskip}\label{eq:convexified_cons}
\end{align}%
Recall that the linearized problem may be infeasible, or the optimal solution to the linearized may no longer be feasible to the dual problem. We alleviate this issue as follows.
First, we add penalty variables to the linearized constraints to ensure that the dual problem is always feasible, and we penalize violations by adding these variables to the objective.
Second, we include \emph{trust regions} around the previous solution to ensure that we do not deviate too much from this solution.
Finally, we incorporate a \emph{robust verification} step to ensure that the obtained policy satisfies the specification.
This verification step is exact and alleviates the potential approximation errors that may arise in the linearization.


\paragraph{Penalty variables.}
We add a nonnegative penalty variable $k_{s}$ for all $ S_{\mathrm{a}}$ to the constraints in~\eqref{eq:convexified_cons}, which yields:
\begin{align}
\forall s \in S_{\mathrm{a}}, r_s \leq  k_s+\!\!\!\sum_{\act \in \Act(s)}  \! \! \! \sched_{s,\act} \!\cdot \!\Big(\rew(s,\act)\! + \!\! \sum_{s' \in S}h_{\textrm{aff}}(s,\act,s')\Big)\raisetag{0.65\baselineskip}
\end{align}
When $k_s$ is sufficiently large, these constraints always allow a feasible solution. We also add a penalty variable to $r_{s_I}\geq \kappa$. 

\paragraph{Trust regions.}
We use trust regions by adding the following set of constraints to the resulting linearized problem:
    \begin{align} 
    &\forall s\in  S,	\quad \nicefrac{\hat{r}_{s}}{\delta'} \leq r_{s}\leq \hat{r}_{s}\cdot\delta',\\
	&\forall s\in  S_{\mathrm{a}},\;\forall \act \in \Act(s),	\quad \nicefrac{\hat{\sched}_{s,\act}}{\delta'}\leq\sched_{s,\act}\leq\hat{\sched}_{s,\act}\cdot\delta',
\end{align}
where $\delta'=\delta+1$ and $\delta>0$ is the size of the trust region, which restricts the set of feasible policies, and $\hat{r}_s$ and $\hat{\sched}_{s,\act}$ denotes the value of the reward and policy variables that are used for linearization.



\paragraph{Extended LP.}
Combining these steps, we now state the resulting finite LP---for some fixed but arbitrary assignment to $\hat{r}_s$ and $\hat{\sched}_{s,\act}$ in the definition of $h_{\textrm{aff}}$, 
penalty parameter $\tau > 0$ and a trust region $\delta > 0$: %
\begin{flalign}
&\begin{aligned}
\text{maximize} \quad   & r_{s_I} - \tau \sum\nolimits_{s \in S_{\textrm{n}}} k_s
\end{aligned} \label{LP:obj}\\
&\begin{aligned}
\text{subject to} \quad
\,\,  r_{s_I} +k_{s_I}\geq \kappa, \;\; \forall s \in G, \quad & r_s =0,
%
\end{aligned} \label{LP:constraint:6}\\
&\begin{aligned} 
\forall s \in S, \quad  & \sum\nolimits_{\act \in \Act(s)} \sched_{s,\act} = 1,
\end{aligned} \label{LP:constraint:3}\\
    &\begin{aligned}  
    \forall s, s' \in S \text{ s.t.\ } \ObsFun(s) = \ObsFun(s'),\forall \alpha \in \Act(s),\;\sigma_{s,\alpha} = \sched_{s',\alpha},
    \end{aligned}\label{LP:constraint:4}\\
&\begin{aligned} 
&\forall s \in S_{\mathrm{a}}, r_s \leq  k_s+\!\!\!\sum_{\act \in \Act(s)}  \! \! \! \sched_{s,\act} \!\cdot \!\Big(\rew(s,\act)\! + \!\! \sum_{s' \in S}h_{\textrm{aff}}(s,\act,s')\Big),\label{LP:convexified_cons}
\end{aligned}\raisetag{0.65\baselineskip}\\
    &\begin{aligned} 
&\forall s \in S_{\mathrm{u}},\,  r_s \leq R(s)+\mu^{\top}_s g_s,\;
 C^\top_s\mu_s+q=0, \;\mu_s \geq 0,
\end{aligned}\label{LP:simple-uncertain-finite}\raisetag{0.15\baselineskip}\\
    &\begin{aligned}  
			 \forall s\in  S,	\quad \nicefrac{\hat{r}_{s}}{\delta'}\leq r_{s}\leq \hat{r}_{s}\cdot\delta',
			 	\end{aligned}\\
			 	    &\begin{aligned}  
		\forall s\in  S_{\mathrm{a}},\;\forall \act \in \Act(s),	\quad \nicefrac{\hat{\sched}_{s,\act}}{\delta'}\leq\sched_{s,\act}\leq\hat{\sched}_{s,\act}\cdot\delta'.\label{LP:trust_region}
			 	\end{aligned}
\end{flalign}

\renewcommand\algorithmicindent{1em}

\begin{algorithm}[t]
	\begin{algorithmic}[1]
	    \Statex \textbf{Input}: uPOMDP, specification with $\kappa$, $\gamma > 1$, $\omega>0$
	    \Statex \textbf{Initialize}: trust region $\delta$, weight $\tau$, policy $\sched$, $\beta_\text{old}=0$. 
		\While{$\delta>\omega$}
		\State \emph{Verify} the uncertain MC with policy $\sched$:\Comment{\textbf{Step 1}}
		\State \emph{Extract} reward variables ${r}_{s}$, objective value $\beta$.
		\If {$\lambda< \beta$} \Comment{\textbf{Step 2}}
		\State \textbf{return} the policy $\sched$ \Comment{Found solution}
		\ElsIf {$\beta> \beta_\text{old}$} 
		\State $\forall s: \hat{r}_s \leftarrow r_s$, $\hat{\sched} \leftarrow \sched$, $\beta_\text{old} \leftarrow \beta$\Comment{Accept iteration}
		\State $\delta \leftarrow\delta \cdot \gamma$ \Comment{Extend trust region}
		\Else
		\State $\delta \leftarrow \delta / \gamma$ \Comment{Reject iteration, reduce trust region}
		\EndIf
		\State \emph{Linearize}~\eqref{NLP:simple-nondet} around $\langle\hat{\sched}, \hat{r}_s\rangle$ \Comment{\textbf{Step 3}}
		\State \emph{Solve} the resulting LP \Comment{ see \eqref{LP:obj}--\eqref{LP:trust_region}}
		\State \emph{Extract}  optimal solution for policy $\sched$
		\EndWhile
		\State \textbf{return} the policy $\sched$
		
		\caption{Sequential convex programming with trust region for solving uncertain POMDPs} 
		\label{alg:scp}
	\end{algorithmic}
\end{algorithm}

\paragraph{Complete algorithm.}
We detail our SCP method in Algorithm~\ref{alg:scp}. 
Its basic idea is to find a solution to the LP  in \eqref{LP:obj}--\eqref{LP:trust_region} such that this solution is feasible to the nonconvex optimization problem with the objective~\eqref{NLP:obj} and the constraints \eqref{NLP:constraint:2}--\eqref{NLP:simple-nondet} and~\eqref{NLP:simple-uncertain-finite}.
We do this by an iterative procedure.
We start with an initial guess of a policy and a trust region radius with $\delta>0$, and (Step 1) we verify the uncertain MC that combines the uPOMDP and the fixed policy $\sched$ to obtain the resulting values of the reward variables $r_{s}$ and an objective value $\beta$. This step is not present in existing SCP methods, and this is our third improvement to ensure that the returned policy robustly satisfies the specification.
If the uncertain MC indeed satisfies the specification (Step 2 -- can be skipped in the first iteration), we return the policy $\sched$.
Otherwise, we check whether the obtained expected reward $\beta$ is improved compared to $\beta_\text{old}$.
In this case, we accept the solution and enlarge the trust region by multiplying $\delta$ with $\gamma>1$.
If not, we reject the solution and contract the trust region by $\gamma$, and resolve the linear problem at the previous solution.
Then (Step 3) we solve the LP~\eqref{LP:obj}--\eqref{LP:trust_region} with the current parameters. We linearize around previous policy variables $\hat{\sched}_{s,\act}$ and reward variables $\hat{r}_s$, and solve with parameters $\delta$ and $\gamma$ to get an optimal solution.
We iterate this procedure until a feasible solution is found, or the radius of the trust region is below a threshold $\omega>0$.
If the trust region size is below $\omega$, we return the policy $\sched$, even though it does not satisfy the expected reward specification. 
In such cases, we can run Algorithm~\ref{alg:scp} with a different initial assignment.

%% file: experiments.tex
\section{Numerical Examples}\label{sec:experiments}
We evaluate the new SCP-based approach to solve the uPOMDP problem on two case studies. Additionally, we compare with~\cite{suilen2020robust}.

\paragraph{Setting.} We use the verification tool Storm $1.6.2$~\cite{DBLP:conf/cav/DehnertJK017} to build uPOMDPs.
The experiments were performed on an Intel
Core i9-9900u 2.50 GHz CPU and 64 GB of RAM with
Gurobi~\cite{gurobi} 9.0 as the LP solver and Storm's robust verification algorithm. We initialize $\hat{\sched}$ to be uniform over all actions.
The algorithm parameters are $\tau=10^4, \delta=1.5, \gamma=1.5,$ and $ \omega=10^{-4}$.



\subsection{Spacecraft Motion Planning}

\begin{figure*}[t]
	\centering
	\begin{subfigure}[t]{0.32\textwidth}
		\centering
	\begin{tikzpicture}
	\begin{axis}[
	xlabel={Time elapsed (s)},
	ylabel style={at={(axis description cs:0.15,.5)},anchor=south}, 
	ylabel=Probability,
	grid=major,
	height=3.7cm,
	width=5.7cm,
	minor y tick num=1,
	ymin=0.6,
	ymax=1.02,
	xmin=-10,
	xmax=1850,
	label style={font=\bf\footnotesize},
	legend cell align={left},
	legend style={at={(1.06,0.4)},fill opacity=1, draw opacity=1,legend columns=2, text opacity=1, draw=white!0.0!black,scale=0.40,font=\scriptsize\selectfont},]

	\addplot  [blue, mark=*,mark size=1.4pt] table [x=x,y=y] {
		y x	
0.049039571623483304 10.711297988891602
0.29006751659217567 19.434131622314453
0.4178584125790903 27.40505886077881
0.5701773211828653 35.352859020233154
0.6193305422382034 43.690833568573
0.6449329565813415 51.482855796813965
0.6454069116325971 60.19274663925171
0.6459158989711008 68.2448468208313
0.6460488469712017 84.483323097229
0.6466405275522096 92.29898166656494
0.6477928357037299 116.75640201568604
0.6492193255111742 124.75153636932373
0.6500091534612612 132.43387365341187
0.6545737501807823 140.2068133354187
0.655460988676118 148.15925121307373
0.6561721123483317 156.03108835220337
0.6579008712030416 163.8514757156372
0.6587317990735533 171.65570163726807
0.659631218082939 179.5492935180664
0.6603315364078041 187.2772617340088
0.6608087680596133 195.088387966156
0.6618502325680752 202.84278964996338
0.6625319599511572 210.56230545043945
0.6635210811609239 218.4358582496643
0.6644115707942341 226.1950821876526
0.6653435854735374 233.93798160552979
0.6663204452862146 241.71958112716675
0.6673351782921854 249.45788764953613
0.6684121075510412 257.19567823410034
0.6695105090921981 264.9880757331848
0.6697599259379643 272.7092423439026
0.6698367258305401 280.52260541915894
0.6699734747246805 288.27314949035645
0.6701693156050679 296.03113412857056
0.6702900754805394 303.7886872291565
0.6703886989922115 311.53833866119385
0.6704647269720675 319.27480459213257
0.6705263997024327 328.874165058136
0.6705966330253424 336.9059190750122
0.6711171616724766 344.7869243621826
0.6714052172053669 353.5109477043152
0.6715926447165566 361.21938037872314
0.6718097818476151 369.0329022407532
0.6718744715478829 376.9659252166748
0.6719421842542975 384.7050414085388
0.6719899811289687 392.99087285995483
0.6720460048281254 400.733925819397
0.6720610545455126 408.454469203949
0.6721309747108847 416.39964628219604
0.6721736400492672 424.1578788757324
0.6722268763618869 432.2336287498474
0.6722437965192982 439.97775888442993
0.6723101496329229 448.063006401062
0.6723500616568321 456.1874861717224
0.6723992500107591 463.90454149246216
0.6724291472446836 472.1203064918518
0.6724910133201208 479.993727684021
0.6725381210741439 488.07802391052246
0.6725991159566351 496.15427827835083
0.6726199388851438 503.8511915206909
0.6727017148345883 512.4509544372559
0.6727241231110622 520.2588143348694
0.6728105058985351 528.0053806304932
0.6728375176921247 536.3555860519409
0.6729377171402845 544.4702668190002
0.6729644275213047 553.5133991241455
0.6730565612496604 561.7078437805176
0.6731018407802638 569.448007106781
0.673201829144606 578.254515171051
0.6732445531317338 585.9775929450989
0.6733253039012385 593.7878217697144
0.6735260933170288 602.2764186859131
0.6738388300975086 609.9734787940979
0.6739212853869386 618.5512742996216
0.6739635359386007 627.0615539550781
0.6739852367923694 634.760929107666
0.6740023798635644 643.3142876625061
0.6756498680517345 651.6597199440002
0.67601594953568 659.3943767547607
0.679116334768008 717.0251741409302
0.6819288569990112 732.3523736000061
0.682943112696197 748.6390223503113
0.6833334426217117 765.1252608299255
0.684331253043988 781.7439932823181
0.6864940468231651 797.7269897460938
0.6900724202925355 806.3954725265503
0.6928328845129772 814.1844892501831
0.7019995520193432 838.1243858337402
0.7065097232490732 870.5081968307495
	};
	\addlegendentry{S1}
	\addplot  [red, mark=diamond*,mark size=1.8pt] table [x=x,y=y] {
		y x	
0.05413447347532223 133.98240637779236
0.3794383259367943 223.98193666934966
0.5626918633469786 316.9117871046066
0.7218634828581979 409.6535421133041
0.7904371188194484 499.6889374017715
0.8148947244663842 591.711310505867
0.8176503048564807 682.9063773870467
0.8216547910174156 774.8251373291015
0.8309718518560779 865.6874832391738
0.8401928296895864 955.8014863491057
0.8537485287283277 1044.429297041893
0.8879143883327192 1133.116036081314
0.890169618722498 1221.9651698589325
0.890586941103829 1312.0293303489684
0.8908570836765372 1402.1540159940719
0.8915906920751421 1758.7958012104034
	};
	\addlegendentry{S2}
		\addplot  [purple, mark=square*,mark size=1.8pt] table [x=x,y=y] {
		y x	
		0.018746683999039386 39.835777902603155
		0.2804859894997502 76.90939741134645
		0.8502593281232295 109.35361824035645
		0.9485340085308535 139.2403377056122
		0.9752895881241137 197.69676609039308
		0.9773631843872124 253.72636752128602
		0.977709310470678 317.2665324687958
		0.9779203677665795 358.8918428897858
		0.9780036651667804 474.67273745536806
		0.9780467095331388 533.5964976787568
		0.9781607827408987 578.6007007598878
		0.9781669239263358 652.526451921463
		0.9782087171987026 696.9452451229096
		0.978245693517658 734.627688074112
		0.9782787930818796 771.2425317287446
		0.9783553275961259 806.759736776352
		0.9783699034395057 846.503890323639
		0.9784422523653481 882.3797842502595
		0.9784733675565254 919.1386382579805
		0.9784959187429515 1035.9477636814117
		0.9785068138590157 1072.0156689643861
		0.9785543466283974 1123.3435922145845
		0.9785708777143496 1156.014478445053
		0.9786167946617725 1192.4495698451997
		0.9786455032715607 1235.6947963714601
		0.9787161185128128 1315.1028171539308
		0.9787802916346510 1475.2603812398503
		0.9788132560112039 1681.9286351298753
	};
	\addlegendentry{S3}
	\addplot  [teal, mark=triangle*,line width=0.25mm,mark size=1.6pt] table [x=x,y=y] {
		y x	
		0.019180298668657832 158.79515986442567
		0.32054408770314985 265.3529024600983
		0.7755067657659118 378.8726950645447
		0.9168532086784618 488.13235955238343
		0.9706214827290212 594.2434663295746
		0.9746941795782206 698.0637911319733
		0.9749111549171017 834.1500594615937
		0.9750869683028153 942.9236237049104
		0.9752146532865608 1180.953446006775
		0.9752746738217909 1314.6799641609193
		0.9794275511582295 1446.6178918361666
		0.9794290043026951 1576.4637748241425
		0.9794939465995538 1710.3258999824525
	};
	\addlegendentry{S4}
	\end{axis}
	\end{tikzpicture}
	\caption{Obtained probability of avoiding close encounters between the spacecraft and other objects in the orbit.}
	\label{fig:conv}	
	\end{subfigure}%
	\hfill
	\begin{subfigure}[t]{0.32\textwidth}
		\centering
	\begin{tikzpicture}
	\begin{axis}[
	xlabel={Time elapsed (s)},
	ylabel style={at={(axis description cs:0.15,.5)},anchor=south}, ylabel=Probability,
	grid=major,
	height=3.7cm,
	width=5.7cm,
	minor y tick num=2,
	ymin=-0.05,
	ymax=1.06,
	xmin=-10,
	xmax=520,
	label style={font=\bf\footnotesize},
	legend cell align={left},
	legend style={at={(1.10,0.75)},fill opacity=1, draw opacity=1,legend columns=1, text opacity=1, draw=white!0.0!black,scale=0.40,font=\scriptsize\selectfont},]

	\addplot  [blue, mark=*,mark size=1.4pt] table [x=x,y=y] {
		y x	
0.3704867009957345 13.023569440841676
0.9232503451420928 22.502764534950256
	};
	\addlegendentry{S1N}
		\addplot  [red, mark=diamond*,mark size=1.8pt] table [x=x,y=y] {
		y x	
		0.2584096205538867 151.2073699951172
		0.8962679669230348 247.24049820899964
		0.9751997537294345 348.2771932125092
	};
	\addlegendentry{S2N}
			\addplot  [purple, mark=square*,mark size=1.8pt] table [x=x,y=y] {
		y x	
		0.20175793073476558 39.91316936016083
		0.8689070528797759 69.44899106025696
		0.969379469795861 98.35730473995208
		0.985407755558627 139.5803535461426
	};

	\addlegendentry{S3N}
	\addplot  [teal, mark=triangle*,line width=0.25mm,mark size=1.6pt] table [x=x,y=y] {
		y x	
		0.19618533208508776 147.14273207187654
		0.8588684359542212 247.5888686656952
		0.9679619926349523 347.03889734745024
		0.9827839480246009 481.55952394008636
	};
	\addlegendentry{S4N}
	\addplot  [blue, dashed,mark=*,mark size=1.4pt] table [x=x,y=y] {
		y x	
0.03234006508515113 13.023569440841676
0.31928392367889985 22.502764534950256
	};
	\addplot  [red, dashed, mark=diamond*,mark size=1.4pt] table [x=x,y=y] {
		y x	
0.01199022125435481 151.2073699951172
0.3205196428796581 247.24049820899964
0.6559986134851427 348.2771932125092
	};
	\addplot  [purple,dashed, mark=square*,mark size=1.8pt]table [x=x,y=y] {
	y x	
	0.010169609710746538 39.91316936016083
	0.39730191036162377 69.44899106025696
	0.7641242905048959 98.35730473995208
	0.8295764433720578 139.5803535461426
};
	\addplot  [teal, dashed,mark=triangle*,line width=0.25mm,mark size=1.6pt] table [x=x,y=y] {
	y x	
	0.009946217095591743 147.14273207187654
	0.35198068547817185 247.5888686656952
	0.753541151282774 347.03889734745024
	0.8110592104928093 481.55952394008636
};
	\end{axis}
	\end{tikzpicture}
	\caption{The performance of the policies obtained from the nominal model applied to the uncertain model (dashed lines). }
	\label{fig:conv_nominal}

	\end{subfigure}%
	\hfill
	\begin{subfigure}[t]{0.32\textwidth}
		\centering
	\begin{tikzpicture}
	\begin{axis}[
	xlabel={Time elapsed (s)},
	ylabel style={at={(axis description cs:0.1,.5)},anchor=south},
	ylabel=Expected cost,
	grid=major,
	height=3.7cm,
	width=5.7cm,
	minor y tick num=6,
	ymin=100,
	ymax=1800,
	ymode=log,
	xmin=-10,
	xmax=1850,
yminorticks=true, 
	label style={font=\bf\footnotesize},
	legend cell align={left},
	legend style={at={(1.0,1.0)},fill opacity=1, draw opacity=1,legend columns=-1, text opacity=1, draw=white!0.0!black,scale=0.40,font=\scriptsize\selectfont},]

	\addplot  [blue, mark=*,mark size=1.4pt] table [x=x,y=y] {
		y x	
32506.133550365234 29.804135751724246
10558.928088182884 56.68813633918763
7237.500428373376 84.97404327392579
1430.6303599574685 115.16891927719116
378.74274646457116 144.2296411037445
239.67009337962787 169.75727319717407
225.06486292898268 217.60124645233157
206.4746797400336 264.498583316803
188.55929312476707 286.53440756797795
182.19767495215893 331.64922580718996
181.29797257257522 445.61290154457095
179.74521562883677 491.22406287193303
178.4012726033152 608.401884508133
178.31581635539237 700.8115550994874
	};
	\addlegendentry{S1}
	\addplot  [red, mark=diamond*,mark size=1.8pt] table [x=x,y=y]{
		y x	
32778.077366871854 168.7615791320801
10680.735596499742 320.99860191345215
7527.721684758576 474.21820907592775
1509.8136409192168 623.4490394592285
702.2586758476705 781.4054370880127
520.1845023336585 943.4925811767579
387.7594923480613 1099.5989959716796
210.71940863829795 1257.1645893096925
189.43067199601808 1402.6078914642335
160.66180019442388 1587.5031639099122
153.7709018752707 1748.0170967102051
	};
	\addlegendentry{S2}
	\end{axis}
	\end{tikzpicture}
	\caption{The obtained expected cost of successfully finishing an orbit.}
	\label{fig:conv_rew}	
	\end{subfigure}
	\caption{Computational effort versus the performance of the different policies for the spacecraft motion planning case study.}

	\end{figure*}
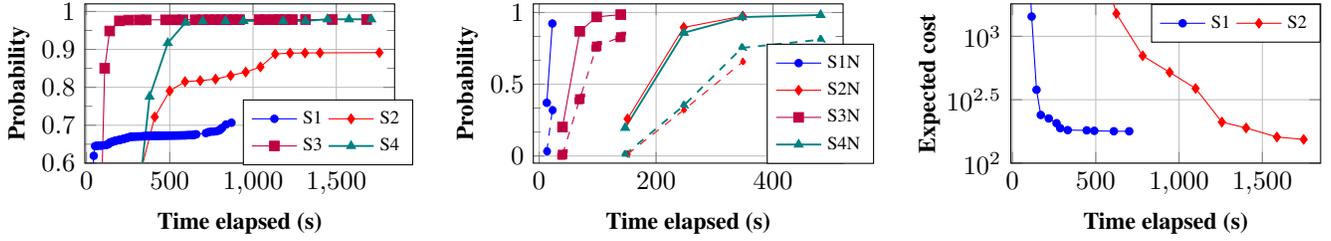

%
This case study considers the robust spacecraft motion planning system~\cite{frey2017constrained,hobbs2020taxonomy}, as mentioned in the introduction.
The spacecraft orbits the earth along a set of predefined natural motion trajectories (NMTs)~\cite{kim2007mission}. 
While the spacecraft follows its current NMT, it does not consume fuel.
Upon an imminent close encounter with other objects in space, the spacecraft may be directed to switch into a nearby NMT at the cost of a certain fuel usage.
We consider two objectives: (1) To maximize the probability of avoiding a close encounter with other objects and (2) to minimize the fuel consumption, both within successfully finishing an orbiting cycle.
Uncertainty enters the problem in the form of potential sensing and actuating errors. 
In particular, there is uncertainty about the spacecraft position, the location of other objects in the current orbit, and the failure rate of switching to a nearby NMT.

\paragraph{Model.}
We encode the problem as a uPOMDP with two-dimensional state variables for the NMT $n~\in~\{1,\ldots,36\}$, and the (discretized) time index $i~\in~\{1,\ldots,I\} $ for a fixed NMT. We use different values of resolution $I$ in the examples. 
Every combination of $\langle n, i\rangle$ defines an associated point in the 3-dimensional space.
The transition model is built as follows.
In each state, there are two types of actions, (1) staying in the current NMT, which increments the time index by $1$, and (2) switching to a different NMT if two locations are close to each other.
More precisely, we allow a transfer between two points in space defined by $\langle n,i\rangle$ and $\langle n',i'\rangle$ if the distance between the two associated points in space is less than 250km. 
A switching action may fail. 
In particular, the spacecraft may transfer to an unintended nearby orbit.
The observation model contains $1440$ different observations of the current locations of the orbit, which give partial information about the current NMT and time index in orbit.
Specifically, for each NMT, we can only observe the location up to an accuracy of $40$ different locations in each orbit. 
The points that are close to each other have the same observation.

\paragraph{Variants.}
We consider three benchmarks. 
S1 is our standard model with a discretized trajectory of $I=200$ time indices.
S2 denotes an extension of S1 where the considered policy is an FSC with $5$ memory states.
S3 uses a higher resolution ($I=600$). 
Finally, S4 is an extension of S3, where the policy is an FSC with $2$ memory states.
In all models, we use an uncertain probability of switching into an intended orbit and correctly locating the objects, given by the intervals $[0.50, 0.95]$. 
The four benchmarks have $36\,048, 349\,480, 108\,000,$ and $342\,750$ states as well as $65\,263, 698\,960, 195\,573,$ and $665\,073$ transitions, respectively.
In this example, the objective is to maximize the probability of avoiding a close encounter with objects in the orbit while successfully finishing a cycle.
\begin{figure*}[t]
\centering
\begin{subfigure}{0.499\textwidth}
\centering
\includegraphics[trim =0 0 0 0, clip, width=0.999\linewidth]{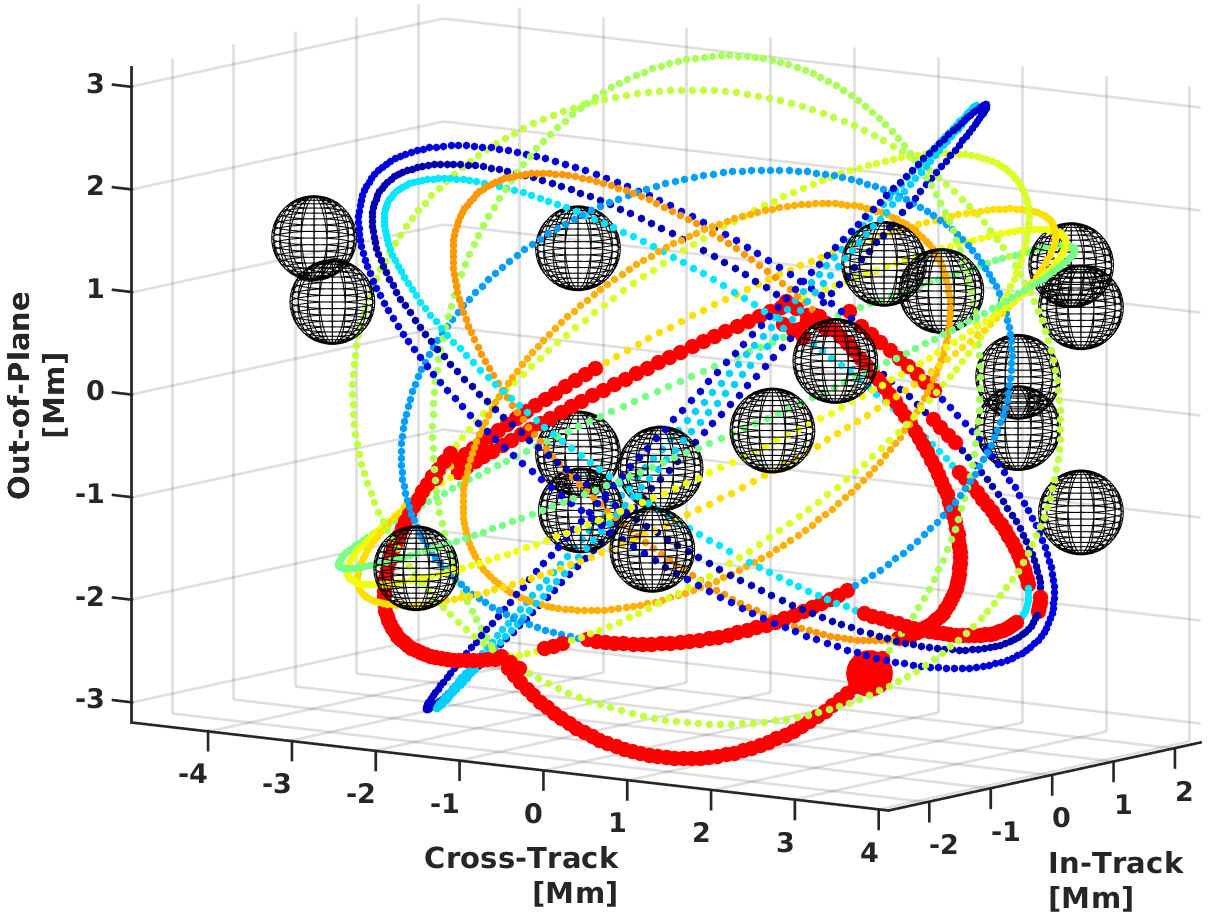}	
	\caption{Trajectory from a memoryless policy.}
	\label{fig:sat1}	
\end{subfigure}%
\hfill
	\begin{subfigure}{0.499\textwidth}
	\centering
	\includegraphics[trim = 0 0 0 0, clip, width=0.999\linewidth]{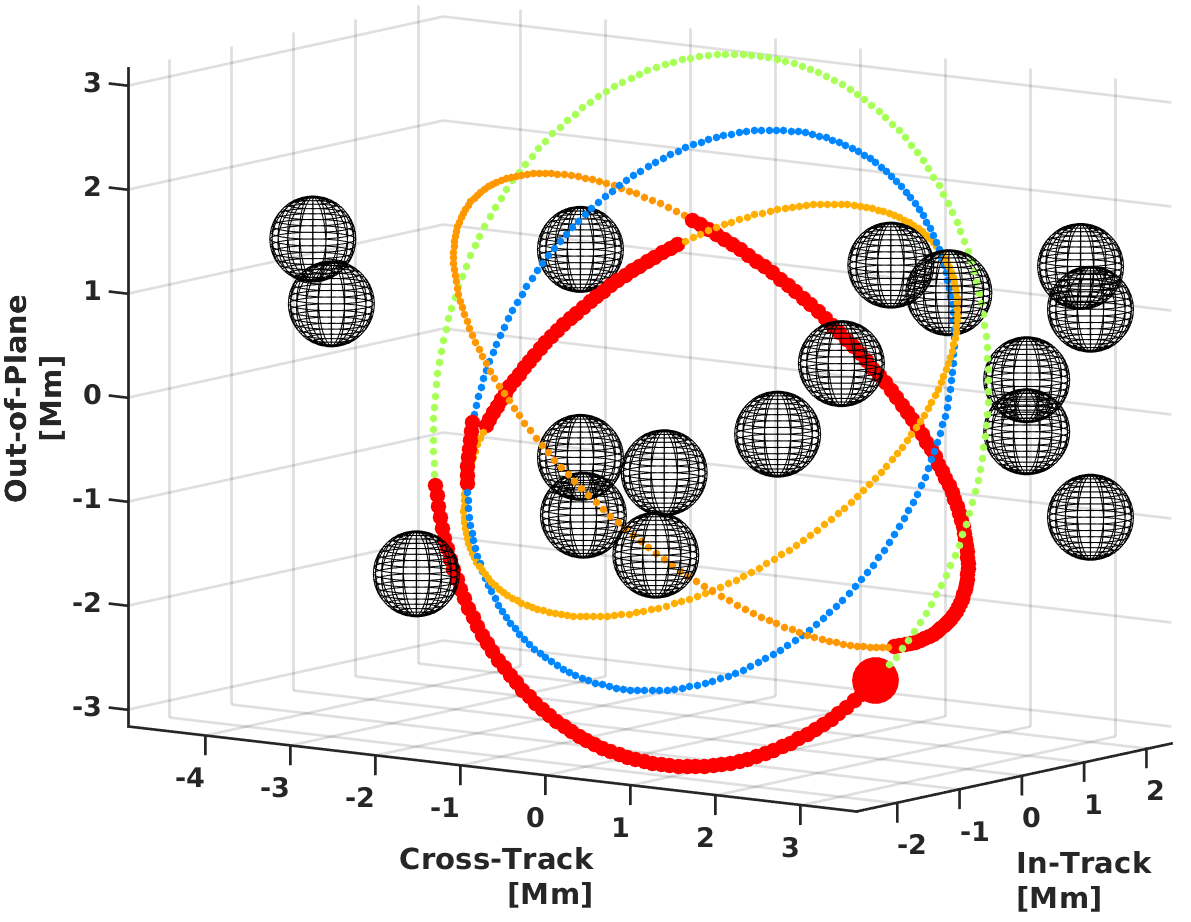}	
	\caption{Trajectory from policy with 5 memory states.}
	\label{fig:sat5}
	\end{subfigure}
	\caption{We show the obtained trajectory from a  policy in red that finishes an orbit around the origin. 
We highlight the initial location by a big red circle.  We depict the other objects by black spheres, and all NMTs that were used as a part of the trajectory.}
	\label{fig:sat}
\end{figure*}

\paragraph{Memory yields best policies.}
Fig.~\ref{fig:conv} shows the convergence of the reachability probabilities for each model, specifically the probability versus the runtime in seconds. 
First, we observe that after 20 minutes of computation, using larger POMDPs that have a higher resolution or memory in the policy yields superior policies. Second, the policy with memory is superior to the policy without memory.
Finally, we observe that larger models indeed require more time per iteration, which means that on the smaller uPOMDP S1, the algorithm converges faster to its optimum. 

\paragraph{Comparing policies.}
Fig.~\ref{fig:sat} shows a comparison of policies and depicts the resulting spacecraft trajectories.
In particular, we show the trajectories from two different policies, a memoryless policy (one memory state) in Fig.~\ref{fig:sat1} -- computed on S1 -- and from a policy with finite memory (five memory states) in Fig.~\ref{fig:sat5} -- computed on S2. 
The trajectory from the memoryless policy switches its orbit $17$ times, whereas the trajectory from the finite-memory policy switches its orbit only $4$ times.
Additionally, the average length of the trajectory with the finite-memory policy is $188$, compared to $375$ for the memoryless one.
These results demonstrate the utility of finite-memory to improve the reachability probability and minimize the number of switches.


%

\paragraph{Robust policies are more robust.}
We demonstrate the utility of computing robust policies against uncertainty in Fig.~\ref{fig:conv_nominal}.
Intuitively, we compute policies on nominal models and use them on uncertain models.
We give results on the nominal transition probabilities of the four considered models, where we ran Algorithm 1 on the nominal models until we reach the optimal value. 
The performance of the policies on the nominal models has  solid lines, and the performance of the policies on the uncertain models has dashed lines.
However, \emph{when we extract these policies and apply them on the uPOMDP, they perform significantly worse and fail to satisfy the objective in each case.}
The results clearly demonstrate the trade-offs between computing policies for nominal and uncertain problems.
In each case, the computation time is roughly an order of magnitude larger. 
Yet, the resulting policies are better: we observe that the probability of a close encounter with another object increases up to $60$ percentage points, if we do not consider the uncertainty in the model.

\paragraph{Expected energy consumption.}
Finally, we consider an example where there is a cost for switching orbits.
The objective is to minimize the cost of successfully finishing a cycle in orbit.
We obtain the cost of each switching action according to the parameters in~\cite{frey2017constrained}.
Additionally, we define the cost of a close encounter to be $10000 N{\cdot}s$ (Newton-second, which is a unit for the impulse and relates to the fuel consumption), a much higher cost than all possible switching actions.
We reduce the uncertainty in the model by setting the previously mentioned intervals for these models to $[0.90, 0.95]$.
In particular, the worst-case probability to correctly detect objects is now higher than before, reducing the probability of close encounters with those objects.
Fig.~\ref{fig:conv_rew} shows the convergence of the costs for each model.
The costs of the resulting policies for models S1 and S2 are $178$ and $153 N{ \cdot }s,$ respectively.
We ran into numerical troubles during the robust verification step for S3 and S4.
Similar to the previous example, the results demonstrate the utility of finite-memory policies to reduce the fuel cost of spacecraft.

\subsection{Aircraft Collision Avoidance}
We consider robust aircraft collision avoidance~\cite{kochenderfer2015decision}.
The objective is to maximize the probability of avoiding a collision with an intruder aircraft while taking into account sensor errors and uncertainty in its future paths.

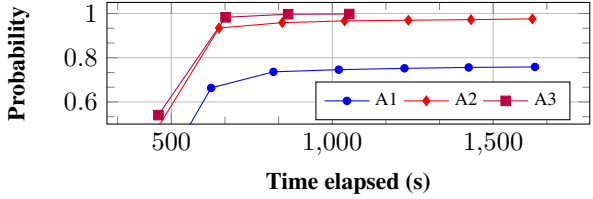
\begin{figure}[t]
	\centering
	\begin{tikzpicture}
	\begin{axis}[
	xlabel={Time elapsed (s)},
	ylabel style={at={(axis description cs:0.05,.5)},anchor=south}, ylabel=Probability,
	grid=major,
	height=3.2cm,
	width=8cm,
 	minor tick num=2,
	ymin=0.5,
	ymax=1.05,
	xmin=300,
	xmax=1800,
	xtick distance=500,
	label style={font=\bf\footnotesize},
	legend cell align={left},
	legend style={at={(0.96,0.35)},fill opacity=1, draw opacity=1,legend columns=-1, text opacity=1, draw=white!0.0!black,scale=0.40,font=\scriptsize\selectfont},]

	\addplot  [blue, mark=*,mark size=1.4pt] table [x=x,y=y] {
		y x	
		0.1792354676216619 431.70761489868164
		0.6635411142693505 623.7513408660889
		0.7361175598853397 817.2198648452759
		0.7459617956458637 1020.2605047225952
		0.7520672987458922 1224.2794704437256
		0.7560751991991371 1423.7664194107056
		0.7581645284942211 1629.903239250183
	};
	\addlegendentry{A1}
	\addplot  [red, mark=diamond*,mark size=1.8pt] table [x=x,y=y]  {
		y x	
		0.4598287361288683 447.94962310791016
		0.9345084077107652 648.064474105835
		0.9585802827016425 844.6246824264526
		0.9665750748798358 1038.2317428588867
		0.9695734345522491 1236.5493659973145
		0.9721849203743244 1430.8338708877563
		0.975655432358234 1621.7335109710693
	};
	\addlegendentry{A2}
		\addplot  [purple, mark=square*,mark size=1.8pt] table [x=x,y=y] {
		y x	
		0.5406739842621497 459.84996604919434
		0.9828024457746893 668.6029233932495
		0.9967512950191275 862.7840204238892
		0.9975238567474634 1052.7180032730103
	};
	\addlegendentry{A3}
	\end{axis}
	\end{tikzpicture}
	\caption{The probability of safely finishing the mission without collision in the aircraft collision example.}
	\label{fig:aircraft}
\end{figure}
\paragraph{Model.}
We consider a two-dimensional example with a state space consisting of (1) the own aircraft position $\langle x, y\rangle$ relative to the intruder, and (2) the relative speed of the intruder relative to the own aircraft, $\langle \dot{x}, \dot{y} \rangle$.
We (grid-)discretize the relative position in 900 cells of $\pm 100\times100$ feet, and the relative speed in 100 cells of  $\dot{x}$ and $\dot{y}$ variables into 10 points between $\pm 1000 \times 1000$ feet/minute (in each direction). We use time steps of $1$ second. 
In each time step, the action choice reflects changing the acceleration of the own aircraft in two dimensions, while the intruder acceleration is chosen probabilistically.
The state is then updated accordingly.
The probabilistic changes by the intruder are given by uncertain intervals $I$, generalizing the model from~\cite{kochenderfer2015decision} where fixed values are assumed.
We vary $I$ below.
Additionally, the probability of the pilot being responsive at a given time is given by the interval $[0.70, 0.90]$.
The own aircraft cannot precisely observe the intruder's position, and the accuracy of the observation is quantized by $\pm 300$ feet.
The model has $476\,009$ states and $866\,889$ transitions.
The specification is to maximize the probability of reaching the target while maintaining a distance of $\pm 400$ feet in either direction.

\paragraph{Variants.}
We consider three ranges for intruder behavior in our example.
The intruder A1 has an uncertain probability of applying an acceleration in one particular direction by the interval ${I = [0.2, 0.8]}$, while we have ${I = [0.3, 0,7]}$ and ${I= [0.4, 0.6]}$ for A2 and A3, respectively.


\paragraph{Results.}
We show the convergence of the method for each intruder in Fig.~\ref{fig:aircraft}.
We compute a locally optimal policy within minutes in each case.
For the three models, the probability of satisfying the specification is $0.997$, $0.976$, and $0.76$, respectively.
Last, we extract the policy to counter intruders within A3 and apply it to the uPOMDP defined for A1, which has more uncertainty than A3.
Then, the resulting reachability probability is only $0.55$, which shows the utility of considering robustness against potentially adversarial intruders.
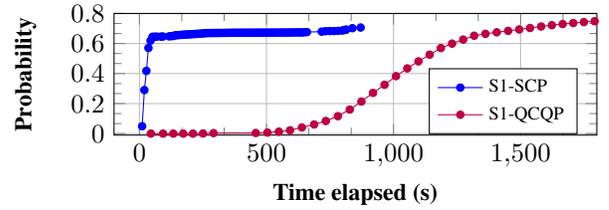
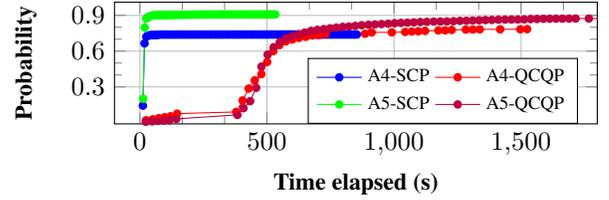
\begin{figure}
\centering
\begin{subfigure}{0.47\textwidth}
	\centering
\begin{tikzpicture}
\begin{axis}[
xlabel={Time elapsed (s)},
ylabel style={at={(axis description cs:0.05,.5)},anchor=south}, ylabel=Probability,
grid=major,
height=3.2cm,
width=8cm,
minor tick num=2,
ymin=-0.01,
ymax=0.8,
xmin=-100,
xmax=1800,
label style={font=\bf\footnotesize},
legend cell align={left},
legend style={at={(0.96,0.55)},fill opacity=1, draw opacity=1,legend columns=1, text opacity=1, draw=white!0.0!black,scale=0.40,font=\scriptsize\selectfont},]

\addplot  [blue, mark=*,mark size=1.4pt] table [x=x,y=y] {
	y x	
0.049039571623483304 10.711297988891602
0.29006751659217567 19.434131622314453
0.4178584125790903 27.40505886077881
0.5701773211828653 35.352859020233154
0.6193305422382034 43.690833568573
0.6449329565813415 51.482855796813965
0.6454069116325971 60.19274663925171
0.6459158989711008 68.2448468208313
0.6460488469712017 84.483323097229
0.6466405275522096 92.29898166656494
0.6477928357037299 116.75640201568604
0.6492193255111742 124.75153636932373
0.6500091534612612 132.43387365341187
0.6545737501807823 140.2068133354187
0.655460988676118 148.15925121307373
0.6561721123483317 156.03108835220337
0.6579008712030416 163.8514757156372
0.6587317990735533 171.65570163726807
0.659631218082939 179.5492935180664
0.6603315364078041 187.2772617340088
0.6608087680596133 195.088387966156
0.6618502325680752 202.84278964996338
0.6625319599511572 210.56230545043945
0.6635210811609239 218.4358582496643
0.6644115707942341 226.1950821876526
0.6653435854735374 233.93798160552979
0.6663204452862146 241.71958112716675
0.6673351782921854 249.45788764953613
0.6684121075510412 257.19567823410034
0.6695105090921981 264.9880757331848
0.6697599259379643 272.7092423439026
0.6698367258305401 280.52260541915894
0.6699734747246805 288.27314949035645
0.6701693156050679 296.03113412857056
0.6702900754805394 303.7886872291565
0.6703886989922115 311.53833866119385
0.6704647269720675 319.27480459213257
0.6705263997024327 328.874165058136
0.6705966330253424 336.9059190750122
0.6711171616724766 344.7869243621826
0.6714052172053669 353.5109477043152
0.6715926447165566 361.21938037872314
0.6718097818476151 369.0329022407532
0.6718744715478829 376.9659252166748
0.6719421842542975 384.7050414085388
0.6719899811289687 392.99087285995483
0.6720460048281254 400.733925819397
0.6720610545455126 408.454469203949
0.6721309747108847 416.39964628219604
0.6721736400492672 424.1578788757324
0.6722268763618869 432.2336287498474
0.6722437965192982 439.97775888442993
0.6723101496329229 448.063006401062
0.6723500616568321 456.1874861717224
0.6723992500107591 463.90454149246216
0.6724291472446836 472.1203064918518
0.6724910133201208 479.993727684021
0.6725381210741439 488.07802391052246
0.6725991159566351 496.15427827835083
0.6726199388851438 503.8511915206909
0.6727017148345883 512.4509544372559
0.6727241231110622 520.2588143348694
0.6728105058985351 528.0053806304932
0.6728375176921247 536.3555860519409
0.6729377171402845 544.4702668190002
0.6729644275213047 553.5133991241455
0.6730565612496604 561.7078437805176
0.6731018407802638 569.448007106781
0.673201829144606 578.254515171051
0.6732445531317338 585.9775929450989
0.6733253039012385 593.7878217697144
0.6735260933170288 602.2764186859131
0.6738388300975086 609.9734787940979
0.6739212853869386 618.5512742996216
0.6739635359386007 627.0615539550781
0.6739852367923694 634.760929107666
0.6740023798635644 643.3142876625061
0.6756498680517345 651.6597199440002
0.67601594953568 659.3943767547607
0.679116334768008 717.0251741409302
0.6819288569990112 732.3523736000061
0.682943112696197 748.6390223503113
0.6833334426217117 765.1252608299255
0.684331253043988 781.7439932823181
0.6864940468231651 797.7269897460938
0.6900724202925355 806.3954725265503
0.6928328845129772 814.1844892501831
0.7019995520193432 838.1243858337402
0.7065097232490732 870.5081968307495
};
\addlegendentry{S1-SCP}
\addplot  [purple, mark=*,mark size=1.4pt] table [x=x,y=y] {
	y x	
7.791005616329295e-06 44.9524040222168
1.3175839470863418e-05 92.96567344665527
1.6738132322753216e-05 133.24579238891602
1.7753128871021633e-05 172.23453426361084
1.809129024241307e-05 210.6789436340332
0.000819672742202926 250.14563751220703
0.003402967622044455 291.30003356933594
0.0041870817842241835 457.98644638061523
0.006899311837346364 503.6237163543701
0.012986870740828563 548.7600688934326
0.022048974439319014 592.7318201065063
0.04105578560698538 639.434944152832
0.06098489876519032 687.6403388977051
0.08406312118432586 733.5107574462891
0.1161220908379329 780.9347305297852
0.16012225157686097 827.7182779312134
0.21351834406843093 873.5817623138428
0.27113183327701623 919.2229919433594
0.32516471616138604 964.5384883880615
0.38278777069504244 1009.8450450897217
0.4344818807917822 1054.0504426956177
0.4811994004378536 1097.9448318481445
0.5260043246931625 1142.141445159912
0.5696006230765193 1186.1417760849
0.5992384249500586 1230.2644538879395
0.626283412572683 1274.220603942871
0.6519029759051469 1317.1874179840088
0.6644944564722368 1361.610852241516
0.674646305398644 1406.2348680496216
0.6842810618534303 1451.001000404358
0.6937706473340419 1494.598183631897
0.7033578243816614 1537.0814266204834
0.7131232178902313 1580.7957935333252
0.721956828945808 1623.0622253417969
0.729415120354144 1665.5061826705933
0.7364313631533971 1707.1099252700806
0.742437873031198 1749.0433292388916
0.7484619398443486 1791.1383848190308
};
\addlegendentry{S1-QCQP}

\end{axis}
\end{tikzpicture}
	\caption{On model S1}
	\label{fig:compare:sat}
\end{subfigure}
\begin{subfigure}{0.47\textwidth}
\centering
\begin{tikzpicture}
\begin{axis}[
	xlabel={Time elapsed (s)},
	ylabel style={at={(axis description cs:0.05,.5)},anchor=south}, ylabel=Probability,
	ytick={0.3,0.6,0.9},
	grid=major,
	height=3.2cm,
	width=8cm,
	minor tick num=2,
	ymin=-0.01,
	ymax=1.01,
	xmin=-100,
	xmax=1800,
	label style={font=\bf\footnotesize},
	legend cell align={left},
	legend style={at={(0.96,0.542)},fill opacity=1, draw opacity=1,legend columns=2, text opacity=1, draw=white!0.0!black,scale=0.40,font=\scriptsize\selectfont},]
	\addplot  [blue, mark=*,mark size=1.4pt] table [x=x,y=y] {
		y x	
0.14277980751042843 11.716261863708496
0.6650057507875299 18.305700302124023
0.7199690572550015 25.402552604675293
0.7275488446499833 31.51517105102539
0.7336235022426507 37.37662124633789
0.7341740737204953 43.19784069061279
0.7344182736430602 48.99394512176514
0.7348002651895622 55.63616752624512
0.7350500925402493 61.39150619506836
0.7352782403807541 67.20478630065918
0.7382376862344919 72.99261856079102
0.738927142906801 78.78944778442383
0.73893373816846 85.04076862335205
0.7389398264130729 91.45725536346436
0.738945943559844 97.28072166442871
0.7389658300792253 103.0751485824585
0.7389810223352201 108.93658542633057
0.7389898425746225 114.7795581817627
0.7389947390524443 121.38039588928223
0.7390042764009802 127.2278881072998
0.7390098207497592 133.0594310760498
0.7390151554413688 138.86597156524658
0.7390204953412881 144.7969741821289
0.7390242151882276 151.16588592529297
0.7390472203404217 157.82411003112793
0.7390539141843964 163.78971481323242
0.7390580340523488 169.73353672027588
0.7390619867908307 175.57855415344238
0.7390663761376753 181.61194324493408
0.7390701469461025 188.0435152053833
0.7390748177238972 194.3847942352295
0.7390790082032997 200.5069169998169
0.739081538290322 206.38379287719727
0.7390846506709818 212.22888278961182
0.7390915139495761 218.0976324081421
0.7390965814733058 224.6243553161621
0.7390996090328908 230.44084644317627
0.7391044403989505 236.32924556732178
0.7391074189290313 242.30571746826172
0.7391105549602534 248.28336811065674
0.7391124791136751 254.59103775024414
0.7391194058152293 261.0682668685913
0.739122294404543 267.14522075653076
0.7391251784366609 273.21222400665283
0.7391289674833701 279.3156671524048
0.7391323842620857 285.40129375457764
0.7391364511930587 292.31968688964844
0.739140510601392 298.5872926712036
0.7391427970252558 304.7118806838989
0.7391456566384682 310.85957050323486
0.7391536960608551 316.833101272583
0.7391562497642985 323.3173007965088
0.739158754945286 330.00682258605957
0.7391643646035608 336.32800674438477
0.7391682858617579 342.5005865097046
0.7391826250005002 348.4182987213135
0.7391928879046609 354.4602289199829
0.7391959136927243 361.62971687316895
0.7391980136317644 367.67921924591064
0.7392001716777367 373.77260875701904
0.7392027566012781 379.66754150390625
0.739204822295688 385.70315170288086
0.7392078634287175 392.03671073913574
0.7392114511415723 398.3936176300049
0.7392173610439486 404.4311304092407
0.7392194287435969 410.49367237091064
0.7392207298035939 416.51465606689453
0.7392269430606796 422.48663234710693
0.739231738209886 429.21522331237793
0.7392412741655848 435.0549592971802
0.7392482407940357 440.9297924041748
0.7392559266971042 446.7954912185669
0.7392636114355972 452.5877294540405
0.7392741320838081 459.01563358306885
0.7392789752325062 465.5525312423706
0.7392952335490041 471.37413024902344
0.7393651331778356 477.31921100616455
0.7393672139084455 483.26799488067627
0.7393691549382412 489.09095668792725
0.7393710809623093 495.62917137145996
0.7393735253034578 501.46449089050293
0.7393761129631328 507.31217193603516
0.739378099618683 513.1141939163208
0.7393795790834701 518.9232559204102
0.7393814576765013 525.1235628128052
0.73938569603063 531.8581504821777
0.7393896957981394 537.9450178146362
0.7393923912931131 544.114423751831
0.7393953238006264 550.1910953521729
0.7393994489138699 556.2226085662842
0.739404205104536 562.6437225341797
0.7394094845023831 568.9093027114868
0.7394115508459235 574.9658288955688
0.7394136126541718 580.7945880889893
0.7394149749524858 586.6622409820557
0.7394197067949313 592.5165395736694
0.7394211125992399 598.4624185562134
0.7394235744367925 604.4361209869385
0.7394269834298187 610.4563722610474
0.7394293106354088 616.4701375961304
0.7394328330181076 622.4051170349121
0.7394351652121346 628.270562171936
0.7394383702912574 634.3098258972168
0.7394398927782536 640.2376432418823
0.7394435056099012 646.0949935913086
0.7394460930294205 651.8771848678589
0.7394483321373511 657.8593864440918
0.7394501142670599 663.7706098556519
0.7394535836861665 669.664436340332
0.7394561441783298 675.6432886123657
0.739458689230103 681.5702972412109
0.7394626362840583 687.6601047515869
0.7394647918482757 693.7164421081543
0.7394669026643699 699.808837890625
0.7394692272598731 705.8275871276855
0.7394716627429331 711.9711408615112
0.7394733024586525 718.0807447433472
0.7394763468542447 723.9132165908813
0.7394775476341945 729.7523822784424
0.7394805119609318 735.6753396987915
0.7394835675938997 741.4550437927246
0.7394844766757229 747.4036874771118
0.7394853816062497 753.3655920028687
0.7394882586510874 759.2732210159302
0.7394913419992527 765.2910575866699
0.739494384651895 771.226957321167
0.7394967122375415 777.1361141204834
0.7394999575157976 783.0909337997437
0.7395030063567939 789.0034685134888
0.7395047922038127 794.9373350143433
0.7395055571490035 800.8530540466309
0.7395096604145397 806.8094244003296
0.7395138604105258 812.6603584289551
0.7395155083853263 818.6681795120239
0.739517848588797 824.5543355941772
0.7395198413338868 830.4349908828735
0.7395218502533489 836.3468980789185
0.7395235919371933 842.2655439376831
0.7395243079718499 848.101713180542
0.7395278051621168 853.8252725601196
	};
	\addlegendentry{A4-SCP}
	\addplot  [red, mark=*,mark size=1.4pt] table [x=x,y=y] {
		y x	
0.019959875419330163 24.133018493652344
0.028236114264908775 50.766090393066406
0.03770006351501474 74.16726875305176
0.04561241330769699 97.26156330108643
0.053559734009620365 120.79850387573242
0.07573776604524791 146.5161542892456
0.08998805575410936 378.7579870223999
0.18595345781086306 404.15521717071533
0.2860264828005975 428.06872177124023
0.3550305441902404 451.94237899780273
0.4070673123250904 478.09481716156006
0.5092490743794738 501.0011444091797
0.5995992283376718 524.866865158081
0.6497346803553266 548.9375314712524
0.6813610236459321 573.8648481369019
0.6971593326852139 599.744610786438
0.7108218685041577 626.2516584396362
0.7196310982529304 652.9473810195923
0.7309168638094142 678.4079103469849
0.7338908009383173 704.5585813522339
0.7436803969688245 730.2007608413696
0.7459732876086052 885.9935178756714
0.7571377471365516 911.9151811599731
0.7573822421254358 991.3170776367188
0.7603803998045532 1043.673300743103
0.7641764383761275 1069.7164974212646
0.7666448227117083 1095.5992164611816
0.7694942508335151 1122.163456916809
0.773455253183276 1174.3030652999878
0.7741754944300377 1200.6887502670288
0.777655990795981 1227.2936162948608
0.7804030512415023 1282.5257186889648
0.780760151447016 1310.7033824920654
0.7822037428438846 1337.646068572998
0.7828587851921517 1417.9460487365723
0.7832240283448467 1497.197522163391
0.7842089363244336 1523.9744338989258
	};
	\addlegendentry{A4-QCQP}
		\addplot  [green, mark=*,mark size=1.4pt] table [x=x,y=y] {
		y x	
0.2009130146511356 12.72978401184082
0.7980489125245763 18.8521785736084
0.8732799513558486 24.86857032775879
0.8855220062681867 30.971434593200684
0.8882566741418896 37.06120300292969
0.8972820563485651 43.131513595581055
0.899485485411077 49.216941833496094
0.9003274177800162 55.45215129852295
0.9007519071243711 62.27666759490967
0.9011497117348011 68.7155237197876
0.9015250272370875 74.84133052825928
0.9018965890079381 81.08790683746338
0.9022361113259797 87.32839012145996
0.9025525055682615 93.77774047851562
0.9028582393073978 100.06867218017578
0.903148657190304 106.74028873443604
0.9034105203837097 113.1802110671997
0.9036631315148953 119.38441944122314
0.9039058218671607 125.70378112792969
0.9041293411675916 132.68488025665283
0.9043356976407254 138.8270959854126
0.9045397334604819 145.05861854553223
0.9047232133386979 151.8735122680664
0.904902514844096 158.4691038131714
0.9050715313460233 164.68220138549805
0.9052286040763234 170.9719638824463
0.9053718484472968 177.18648529052734
0.9055131868667681 183.44861602783203
0.905650912391595 189.6261329650879
0.9057766864228827 196.17807483673096
0.905899855701583 202.71170711517334
0.9060206258272326 208.84182739257812
0.9061320802786024 215.10167789459229
0.906231932397564 221.14573764801025
0.9063256584786236 227.34370231628418
0.9064152363526343 233.65205097198486
0.9065064190976013 240.4307737350464
0.9066041606084773 246.7605972290039
0.9066936408461386 252.9899778366089
0.9068186250521367 258.9634618759155
0.9069528946741231 264.9651041030884
0.9069744637360747 271.0009489059448
0.9069916115815685 277.59685134887695
0.9070003093379226 283.93404483795166
0.9070139973114032 289.96538066864014
0.9070257805541275 295.96592903137207
0.9070367767186657 301.97578144073486
0.9070491186223846 307.9617528915405
0.9070612037314257 314.5950574874878
0.90706884210858 320.7280397415161
0.9070794180754976 326.74428844451904
0.9070865164000241 332.67660427093506
0.9070974904688625 338.64645767211914
0.9071147548816914 344.6093864440918
0.9071263386550115 351.57210063934326
0.9071344112083798 357.7249231338501
0.9071406338433393 363.8552551269531
0.9071460986403592 369.9722604751587
0.9071593382747801 376.1675205230713
0.9071700979838923 382.27666664123535
0.9071788496181814 389.62765312194824
0.9071878259520705 395.9888334274292
0.9072049505232366 402.176815032959
0.907219312373328 408.3483467102051
0.9072308537907616 414.60801792144775
0.907242263967681 421.17536544799805
0.9072537217600468 427.75738048553467
0.9072651103770328 433.98158073425293
0.9072763881068564 439.99461460113525
0.907288312176261 445.99724197387695
0.9072981548302894 452.00367736816406
0.9073094245051027 458.856463432312
0.9073200447890957 464.904109954834
0.9073306974269968 470.96972465515137
0.9073423523789755 476.96667289733887
0.9073620721644288 483.0836572647095
0.9073744135612529 489.6519832611084
0.9073832254899886 496.31661224365234
0.9074037441120814 502.3916988372803
0.9074108062182253 508.4000120162964
0.907432371930811 514.4050712585449
0.9074479743537023 520.4193315505981
0.9074609743541706 527.2045278549194
0.9074736849765571 533.3146562576294
	};
	\addlegendentry{A5-SCP}
	\addplot  [purple, mark=*,mark size=1.4pt] table [x=x,y=y] {
		y x	
0.007701400370734301 23.079015731811523
0.010959509103934924 48.32380676269531
0.014670028636952263 70.90164279937744
0.017761735846184852 93.72980880737305
0.02023401138999499 116.32247734069824
0.03165000903998511 143.2209415435791
0.06524072284054887 383.85399436950684
0.12113392201702505 407.59517192840576
0.18009294701285236 433.9324674606323
0.2926153562816143 457.4283800125122
0.46994503345849226 478.2755308151245
0.5708159090443939 502.5852746963501
0.6331275094492749 526.760274887085
0.6768395263138459 550.7664909362793
0.7066100078341528 575.093560218811
0.7238129123258815 599.3257818222046
0.741069811297334 624.0203971862793
0.7561416947089548 649.1608572006226
0.769988227763632 673.8496465682983
0.7784362710804481 698.4686126708984
0.7845894538639989 723.5365686416626
0.7915020456240393 748.2599401473999
0.7971862046613687 772.9630308151245
0.803042438493926 798.0754899978638
0.8061690739507866 822.735276222229
0.8113791322291259 847.587010383606
0.8173822954627724 872.7723951339722
0.8207772109354998 899.0482797622681
0.8240535589136855 924.9597358703613
0.8269231686141262 950.7794008255005
0.8298538916982562 977.0710544586182
0.8345194253145063 1002.4813575744629
0.836453461401021 1027.2396306991577
0.8383284741548592 1052.49600315094
0.8411256055597243 1077.6750078201294
0.8439566862419906 1103.3086042404175
0.8454509015968975 1129.294023513794
0.8474855715165184 1156.1591033935547
0.8481711450464332 1182.3933849334717
0.8493023229287032 1208.159369468689
0.8509609290474455 1233.9436416625977
0.8533323751501983 1259.3256902694702
0.8544262924576481 1284.534436225891
0.8563386917473989 1309.7542161941528
0.8589118285678375 1335.0941123962402
0.8603363993049378 1359.972933769226
0.8616024015747823 1385.285717010498
0.8635433959767148 1411.0874452590942
0.8642773051184134 1437.0175590515137
0.8651418572654132 1462.4416589736938
0.8652375807505182 1487.2781429290771
0.8663143977390432 1512.5371389389038
0.8687840655928064 1537.8784761428833
0.8694364616594086 1562.9017362594604
0.8712303036192709 1587.9726867675781
0.871597991555277 1612.7903261184692
0.8719437771177838 1637.7662267684937
0.8719855113797537 1662.8602647781372
0.8725976674422379 1688.7916765213013
0.8733027376561459 1714.7528800964355
0.8733134995834744 1766.7916955947876
0.8734628761160171 1817.4502115249634
	};
	\addlegendentry{A5-QCQP}
\end{axis}
\end{tikzpicture}
	\caption{On smaller models A4 and A5.}
	\label{fig:compare:air}
\end{subfigure}
\caption{Comparing SCP with QCQP.}	
\end{figure}
\subsection{Comparing Performance}
We compare our approach (SCP) with an implementation of \cite{suilen2020robust}, based on an exponentially large quadratically constrained quadratic program (QCQP). 

Let us first revisit the spacecraft problem. 
The QCQP method runs out of memory on all models except S1. 
For S1, QCQP indeed has a small advantage over SCP, cf.\ Fig.~\ref{fig:compare:sat}. 
We remark that the policies computed with SCP on S2, S3, and S4 are superior to the policy computed by QCQP on S1.

For the airplane problem, the QCQP again runs out of memory on A1, A2, and A3. 
For comparison, we construct models A4 and A5 by a coarser discretization (and varying uncertainty). 
The models have $18\, 718$ states, and are smaller versions of A1 and A2.
The performance is plotted in Fig.~\ref{fig:compare:air}. 
For A4, QCQP yields a slightly better policy but takes more time, whereas, on A5, SCP finds a better policy faster.

We conclude that QCQP from \cite{suilen2020robust} is comparable on small models, but does not scale to large models.

%% file: conclusion.tex
\section{Conclusion}\label{sec:conclusion}
We presented a new approach to computing robust policies for uncertain POMDPs. 
The experiments showed that we are able to apply our method based on convex optimization on well-known benchmarks with varying levels of uncertainty.
Future work will involve incorporating learning-based techniques.


%

%% file: main.bbl
\begin{thebibliography}{32}
\providecommand{\natexlab}[1]{#1}
\providecommand{\url}[1]{\texttt{#1}}
\providecommand{\urlprefix}{URL }
\expandafter\ifx\csname urlstyle\endcsname\relax
  \providecommand{\doi}[1]{doi:\discretionary{}{}{}#1}\else
  \providecommand{\doi}{doi:\discretionary{}{}{}\begingroup
  \urlstyle{rm}\Url}\fi

\bibitem[{Ahmadi et~al.(2018)Ahmadi, Cubuktepe, Jansen, and
  Topcu}]{ahmadi2018verification}
Ahmadi, M.; Cubuktepe, M.; Jansen, N.; and Topcu, U. 2018.
\newblock {Verification of Uncertain POMDPs Using Barrier Certificates}.
\newblock In \emph{Allerton}, 115--122. IEEE.

\bibitem[{Alizadeh and Goldfarb(2003)}]{alizadeh2003second}
Alizadeh, F.; and Goldfarb, D. 2003.
\newblock {Second-Order Cone Programming}.
\newblock \emph{Math Program.} 95(1): 3--51.

\bibitem[{Amato, Bernstein, and Zilberstein(2010)}]{amato2010optimizing}
Amato, C.; Bernstein, D.~S.; and Zilberstein, S. 2010.
\newblock {Optimizing Fixed-size Stochastic Controllers for {POMDPs} and
  Decentralized {POMDPs}}.
\newblock \emph{AAMAS} 21(3): 293--320.

\bibitem[{Ben-Tal, El~Ghaoui, and Nemirovski(2009)}]{ben2009robust}
Ben-Tal, A.; El~Ghaoui, L.; and Nemirovski, A. 2009.
\newblock \emph{{Robust Optimization}}, volume~28.
\newblock Princeton University Press.

\bibitem[{Bertsimas, Brown, and Caramanis(2011)}]{bertsimas2011theory}
Bertsimas, D.; Brown, D.~B.; and Caramanis, C. 2011.
\newblock {Theory and Applications of Robust Optimization}.
\newblock \emph{SIAM review} 53(3): 464--501.

\bibitem[{Burns and Brock(2007)}]{burns2007sampling}
Burns, B.; and Brock, O. 2007.
\newblock {Sampling-Based Motion Planning with Sensing Uncertainty}.
\newblock In \emph{ICRA}, 3313--3318. IEEE.

\bibitem[{Chamie and Mostafa(2018)}]{DBLP:conf/cdc/ChamieM18}
Chamie, M.~E.; and Mostafa, H. 2018.
\newblock {Robust Action Selection in Partially Observable Markov Decision
  Processes with Model Uncertainty}.
\newblock In \emph{{CDC}}, 5586--5591. {IEEE}.

\bibitem[{Chatterjee et~al.(2016)Chatterjee, Chmel{\'\i}k, Gupta, and
  Kanodia}]{ChatterjeeCGK16}
Chatterjee, K.; Chmel{\'\i}k, M.; Gupta, R.; and Kanodia, A. 2016.
\newblock {Optimal Cost Almost-sure Reachability in POMDPs}.
\newblock \emph{Artif. Intell.} 234: 26--48.

\bibitem[{Dehnert et~al.(2017)Dehnert, Junges, Katoen, and
  Volk}]{DBLP:conf/cav/DehnertJK017}
Dehnert, C.; Junges, S.; Katoen, J.; and Volk, M. 2017.
\newblock {A Storm is Coming: {A} Modern Probabilistic Model Checker}.
\newblock In \emph{{CAV} {(2)}}, volume 10427 of \emph{LNCS}, 592--600.
  Springer.

\bibitem[{Frey et~al.(2017)Frey, Petersen, Leve, Kolmanovsky, and
  Girard}]{frey2017constrained}
Frey, G.~R.; Petersen, C.~D.; Leve, F.~A.; Kolmanovsky, I.~V.; and Girard,
  A.~R. 2017.
\newblock {Constrained Spacecraft Relative Motion Planning Exploiting Periodic
  Natural Motion Trajectories and Invariance}.
\newblock \emph{Journal of Guidance, Control, and Dynamics} 40(12): 3100--3115.

\bibitem[{Givan, Leach, and Dean(2000)}]{givan2000bounded}
Givan, R.; Leach, S.; and Dean, T. 2000.
\newblock {Bounded-Parameter Markov Decision Processes}.
\newblock \emph{Artif. Intell.} 122(1-2): 71--109.

\bibitem[{Gurobi~Optimization(2020)}]{gurobi}
Gurobi~Optimization, L. 2020.
\newblock Gurobi Optimizer Reference Manual.
\newblock \urlprefix\url{http://www.gurobi.com}.

\bibitem[{Hahn et~al.(2017)Hahn, Hashemi, Hermanns, Lahijanian, and
  Turrini}]{DBLP:conf/qest/HahnHHLT17}
Hahn, E.~M.; Hashemi, V.; Hermanns, H.; Lahijanian, M.; and Turrini, A. 2017.
\newblock {Multi-Objective Robust Strategy Synthesis for Interval Markov
  Decision Processes}.
\newblock In \emph{{QEST}}, volume 10503 of \emph{LNCS}, 207--223. Springer.

\bibitem[{Hobbs and Feron(2020)}]{hobbs2020taxonomy}
Hobbs, K.~L.; and Feron, E.~M. 2020.
\newblock {A Taxonomy for Aerospace Collision Avoidance with Implications for
  Automation in Space Traffic Management}.
\newblock In \emph{AIAA Scitech 2020 Forum}, 0877.

\bibitem[{Itoh and Nakamura(2007)}]{itoh2007}
Itoh, H.; and Nakamura, K. 2007.
\newblock {Partially Observable Markov Decision Processes with Imprecise
  Parameters}.
\newblock \emph{Artif. Intell.} 171(8): 453 -- 490.

\bibitem[{Junges et~al.(2018)Junges, Jansen, Wimmer, Quatmann, Winterer,
  Katoen, and Becker}]{junges2018finite}
Junges, S.; Jansen, N.; Wimmer, R.; Quatmann, T.; Winterer, L.; Katoen, J.; and
  Becker, B. 2018.
\newblock {Finite-State Controllers of POMDPs using Parameter Synthesis}.
\newblock In \emph{{UAI}}, 519--529.

\bibitem[{Kaelbling, Littman, and Cassandra(1998)}]{kaelbling1998planning}
Kaelbling, L.~P.; Littman, M.~L.; and Cassandra, A.~R. 1998.
\newblock Planning and Acting in Partially Observable Stochastic Domains.
\newblock \emph{Artificial intelligence} 101(1-2): 99--134.

\bibitem[{Kim et~al.(2007)Kim, Shepperd, Norris, Goldberg, and
  Wallace}]{kim2007mission}
Kim, S.~C.; Shepperd, S.~W.; Norris, H.~L.; Goldberg, H.~R.; and Wallace, M.~S.
  2007.
\newblock {Mission Design and Trajectory Analysis for Inspection of a Host
  Spacecraft by a Microsatellite}.
\newblock In \emph{2007 IEEE Aerospace Conference}, 1--23. IEEE.

\bibitem[{Kochenderfer(2015)}]{kochenderfer2015decision}
Kochenderfer, M.~J. 2015.
\newblock \emph{{Decision Making Under Uncertainty: Theory and Application}}.
\newblock MIT press.

\bibitem[{Lobo et~al.(1998)Lobo, Vandenberghe, Boyd, and
  Lebret}]{lobo1998applications}
Lobo, M.~S.; Vandenberghe, L.; Boyd, S.; and Lebret, H. 1998.
\newblock Applications of Second-Order Cone Programming.
\newblock \emph{{Linear Algebra and its Applications}} 284(1-3): 193--228.

\bibitem[{Madani, Hanks, and Condon(1999)}]{MadaniHC99}
Madani, O.; Hanks, S.; and Condon, A. 1999.
\newblock {On the Undecidability of Probabilistic Planning and Infinite-Horizon
  Partially Observable {Markov} Decision Problems}.
\newblock In \emph{AAAI}, 541--548. {AAAI} Press.

\bibitem[{Mao et~al.(2018)Mao, Szmuk, Xu, and Acikmese}]{mao2018successive}
Mao, Y.; Szmuk, M.; Xu, X.; and Acikmese, B. 2018.
\newblock {Successive Convexification: A Superlinearly Convergent Algorithm for
  Non-convex Optimal Control Problems}.
\newblock \emph{arXiv preprint arXiv:1804.06539} .

\bibitem[{Meuleau et~al.(1999)Meuleau, Kim, Kaelbling, and
  Cassandra}]{meuleau1999solving}
Meuleau, N.; Kim, K.-E.; Kaelbling, L.~P.; and Cassandra, A.~R. 1999.
\newblock {Solving {POMDPs} by Searching the Space of Finite Policies}.
\newblock In \emph{UAI}, 417--426.

\bibitem[{Nilim and Ghaoui(2005)}]{DBLP:journals/ior/NilimG05}
Nilim, A.; and Ghaoui, L.~E. 2005.
\newblock {Robust Control of Markov Decision Processes with Uncertain
  Transition Matrices}.
\newblock \emph{Operations Research} 53(5): 780--798.

\bibitem[{Osogami(2015)}]{DBLP:conf/icml/Osogami15}
Osogami, T. 2015.
\newblock {Robust Partially Observable Markov Decision Process}.
\newblock In \emph{{ICML}}, volume~37, 106--115.

\bibitem[{Puggelli et~al.(2013)Puggelli, Li, Sangiovanni{-}Vincentelli, and
  Seshia}]{DBLP:conf/cav/PuggelliLSS13}
Puggelli, A.; Li, W.; Sangiovanni{-}Vincentelli, A.~L.; and Seshia, S.~A. 2013.
\newblock {Polynomial-Time Verification of PCTL Properties of MDPs with Convex
  Uncertainties}.
\newblock In \emph{CAV}, volume 8044 of \emph{LNCS}, 527--542. Springer.

\bibitem[{Puterman(1994)}]{Put94}
Puterman, M.~L. 1994.
\newblock \emph{{{M}arkov} Decision Processes}.
\newblock John Wiley and Sons.

\bibitem[{Suilen et~al.(2020)Suilen, Jansen, Cubuktepe, and
  Topcu}]{suilen2020robust}
Suilen, M.; Jansen, N.; Cubuktepe, M.; and Topcu, U. 2020.
\newblock Robust Policy Synthesis for Uncertain POMDPs via Convex Optimization.
\newblock In \emph{{IJCAI}}, 4113--4120. ijcai.org.

\bibitem[{Vlassis, Littman, and Barber(2012)}]{VlassisLB12}
Vlassis, N.; Littman, M.~L.; and Barber, D. 2012.
\newblock {On the Computational Complexity of Stochastic Controller
  Optimization in POMDPs}.
\newblock \emph{ACM Trans.\ on Computation Theory} 4(4): 12:1--12:8.

\bibitem[{Wiesemann, Kuhn, and Rustem(2013)}]{wiesemann2013robust}
Wiesemann, W.; Kuhn, D.; and Rustem, B. 2013.
\newblock {Robust Markov Decision Processes}.
\newblock \emph{Mathematics of Operations Research} 38(1): 153--183.

\bibitem[{Wolff, Topcu, and Murray(2012)}]{DBLP:conf/cdc/WolffTM12}
Wolff, E.~M.; Topcu, U.; and Murray, R.~M. 2012.
\newblock {Robust Control of Uncertain Markov Decision Processes with Temporal
  Logic Specifications}.
\newblock In \emph{CDC}, 3372--3379. IEEE.

\bibitem[{Yuan(2015)}]{yuan2015recent}
Yuan, Y.-x. 2015.
\newblock {Recent Advances in Trust Region Algorithms}.
\newblock \emph{Mathematical Programming} 151(1): 249--281.

\end{thebibliography}
